\documentclass[onefignum,onetabnum]{siamsials251208}

\usepackage{lipsum}
\usepackage{amsfonts}
\usepackage{graphicx}
\usepackage{epstopdf}
\usepackage{algorithmic}
\ifpdf
  \DeclareGraphicsExtensions{.eps,.pdf,.png,.jpg}
\else
  \DeclareGraphicsExtensions{.eps}
\fi

\usepackage{enumitem}
\setlist[enumerate]{leftmargin=.5in}
\setlist[itemize]{leftmargin=.5in}

\newsiamremark{remark}{Remark}
\newsiamremark{hypothesis}{Hypothesis}
\crefname{hypothesis}{Hypothesis}{Hypotheses}
\newsiamthm{claim}{Claim}
\newsiamremark{fact}{Fact}
\crefname{fact}{Fact}{Facts}

\headers{LIFT}{B. Yang, J. Ying, Y. Wang, P. Zhou and L. Zhang}

\title{LIFT: Latent Flow Matching for Probabilistic Cell Fate Dynamics with Single Cell Snapshots\thanks{Submitted to the editors \date{}.
\funding{L.Z. was supported by the National Natural Science Foundation of China (No. 12225102, T2321001, and 12288101) and the National Key Research and Development Program of China  2024YFA0919500.}}}

\author{Bowen Yang\thanks{School of Mathematical Sciences, Peking University, Beijing, China 
   (\email{yangbowen@stu.pku.edu.cn})}
\and Junda Ying\thanks{Beijing International Center for Mathematical Research, Peking University, Beijing, China}
\and Yuxuan Wang\thanks{Center for Quantitative Biology, Peking University, Beijing, China}
\and Peijie Zhou\thanks{Center for Quantitative Biology, Center for Machine Learning Research, National Engineering Laboratory for Big Data Analysis and Applications, AI for Science Institute, Peking University, Beijing, China.}
\and Lei Zhang\thanks{School of Mathematical Sciences, Beijing International Center for Mathematical Research, Center for Quantitative Biology, Center for Machine Learning Research, Institute for Artificial Intelligence, Peking University, Beijing 100871, China (\email{zhangl@math.pku.edu.cn})}}

\usepackage{amsopn}

\makeatletter
\let\LIFTsuppmaketitle\@maketitle
\makeatother

\ifpdf
\hypersetup{
  pdftitle={LIFT: Latent Flow Matching for Probabilistic Cell Fate Dynamics with Single Cell Snapshots},
  pdfauthor={}
}
\fi

\usepackage{xspace}
\usepackage{mathrsfs}
\usepackage{booktabs}
\usepackage{tabularx}
\usepackage{subcaption}

\begin{document}

\maketitle

\begin{abstract}

Lineage tracing links cell states through clonal ancestry, but does not generally identify unique source-descendant pairings. Learning from such data therefore requires a model of conditional outcomes, rather than only the terminal population distribution. We introduce LIFT, a latent flow framework for probabilistic cell-state generation from lineage-informed couplings. Variational inference encodes alternative outcomes in a latent space, while a conditional prior enables sampling from the initial state alone. A Mean Flow decoder predicts future cell state directly, requiring one decoder evaluation per sampling, which makes it convenient to predict cell fate through large scale Monte-Carlo sampling. This construction combines a representation of one-to-many relationships with efficient sampling and an analysis of conditional prediction error. Experiments on synthetic and real datasets demonstrate LIFT's ability to recover both population distribution and cell fate probabilities.

\end{abstract}
\begin{relevance}

Predicting differentiation and reprogramming requires estimating both which outcomes are possible and how their probabilities depend on the observed initial state. We evaluate LIFT on lineage-traced mouse hematopoiesis and fibroblast reprogramming using transcriptomic state representations. Lineage-informed training improves fate-probability prediction relative to geometric coupling, while the hematopoietic application illustrates graded fate bias in regions with mixed clonal outcomes. Synthetic branching experiments further demonstrate accurate population reconstruction and branching probabilities at low sampling cost.

\end{relevance}
\begin{mathcontent}

We formulate barcode resolved observations as measures linked by an unknown true lineage-compatible coupling, distinguish this target from the empirical coupling used for training, and quantify reconstruction error via conditional Wasserstein discrepancies. The irreducible error of deterministic endpoint maps is characterized by the true conditional variance. We establish reconstruction estimates for flow matching and Mean Flow decoders and bound the total conditional error by coupling discrepancy, decoder reconstruction, and mismatch between the aggregated posterior and sampling prior. Under regularity and bounded Gaussian prior scale assumptions, the latter terms are controlled by the training objectives. Boundary sensitive probability estimates and Monte Carlo concentration bounds then connect conditional reconstruction to the accuracy of estimated fate proportions, separating model error from finite-sampling uncertainty.

\end{mathcontent}



\section{Introduction}
\label{sec:introduction}

Understanding how an early cell state constrains its future fate is a
central problem in developmental biology and cellular reprogramming. Single-cell RNA
sequencing provides detailed molecular snapshots, but its destructive
nature generally prevents repeated measurements of the same cell.
Trajectory inference must therefore reconstruct developmental
relationships from population snapshots and additional assumptions
\cite{Saelens2019}. Lineage-tracing technologies provide an experimental source of such information. Methods such as CellTagging and LARRY recover
heritable barcodes together with transcriptomic states, linking clonal
ancestry to subsequent outcomes \cite{Biddy2018,Weinreb2020}. These
observations motivate models that use lineage information to predict not
only a likely terminal distribution, but also the probabilities of cellular fates.

A shared barcode does not, however, generally determine a unique pairing
between individual cells observed at different times. When several
source and terminal cells belong to the same clone, their molecular
correspondence remains only partially specified. We represent the true
endpoint relationship by an unknown coupling $\pi_\star$ and distinguish
it from the lineage-informed coupling $\pi$ constructed for training.
The prediction target is the conditional law
\[
K_\star(x,A)
=
\mathbb P_{\pi_\star}(Y\in A\mid X=x),
\]
where $x$ is the observed initial state and $A$ is a region of terminal
state space. Different source-descendant relationships can have identical marginals but imply different fate probabilities. Recovering this law requires more than matching the
endpoint populations. 

Probabilistic models of cellular differentiation range from fate mapping on observed states to continuous descriptions of population dynamics. Palantir estimates terminal fate probabilities from transitions on the transcriptomic manifold \cite{Setty2019}, while CellRank 2 integrates multiple sources of directional and temporal information through Markov state modeling \cite{Weiler2024}. Waddington-OT estimates probabilistic couplings between successive populations while accounting for growth and death \cite{Schiebinger2019}, and PRESCIENT learns stochastic cell-state dynamics from expression time series \cite{Yeo2021}. These approaches already represent heterogeneous outcomes and, in several cases, provide fate probabilities. Without explicit lineage constraints, the inferred fate relationships depend on assumptions or constraints used by each method, but they need not coincide with those supported by lineage observations.

Explicit incorporation of lineage information provides additional constraints on these relationships. LineageOT uses lineage trees to inform ancestral-state estimation and transport coupling \cite{Forrow2021}; CoSpar reconstructs transition maps from sparse clonal observations using sparsity and local coherence \cite{Wang2022}; and moslin combines expression similarity with lineage geometry through fused Gromov--Wasserstein transport \cite{Lange2024Moslin}. LineageVAE further uses variational latent dynamics to reconstruct historical states toward unobserved common progenitors \cite{Majima2024LineageVAE}. These studies establish lineage observations as valuable information for both transition inference and state reconstruction. Nevertheless, an endpoint transition map does not uniquely specify intermediate paths. Estimating a lineage-informed coupling and constructing a continuous generator that retains its conditional structure are thus separate, complementary tasks.

Generative modeling provides tools for constructing continuous conditional models. Flow matching learns time-dependent velocity fields
without numerical integration during training \cite{Lipman2023FM,Tong2024CFM}, and Mean Flow addresses sampling cost by learning
interval-average velocities for one-step generation
\cite{Geng2025MeanFlow},
but a deterministic flow initialized solely at the observed cell state produces only one outcome from that state. Augmented Bridge Matching addresses
coupling preservation by retaining the initial sample in the conditioning
\cite{DeBortoli2023ABM}, while Variational Rectified Flow Matching uses
latent variables to represent multimodal velocity directions
\cite{Guo2025VRFM}. For lineage-informed applications, the remaining task is to combine
one-to-many conditional representation with direct time-dependent
sampling. This combination motivates the LIFT framework
developed below.

We introduce LIFT, a latent flow framework that combines conditional
representation, efficient generation, and error control for this task.
Building on variational inference and conditional latent-variable
modeling \cite{Kingma2014VAE,Sohn2015CVAE}, an encoder infers latent
codes from training pairs, while a conditional prior predicts their
distribution from the initial state alone. We formulate Flow Matching and Mean
Flow reconstruction surrogates and use the Mean Flow decoder for direct
generation. Its average velocity parameterization
\cite{Geng2025MeanFlow} allows only one decoder evaluation per sampling, and keeping the latent code
fixed across time defines a continuous generated path. The coupling can
incorporate geometric or clonal information without changing the
sampling architecture.

Our theoretical analysis makes the conditional prediction objective explicit. We characterize the best deterministic approximation by its irreducible conditional-variance error and establish reconstruction bounds for both FM and Mean Flow decoders. Under the stated regularity and prior-scale assumptions, the total conditional Wasserstein error is bounded by the discrepancy between $\pi$ and $\pi_\star$, reconstruction under encoder samples, and mismatch between the aggregated posterior and the sampling prior. We further translate this bound into fate-probability estimates, including the effects of classification boundaries and finite Monte Carlo sampling. These results concern a fitted model and do not identify the true coupling or a unique intermediate biological trajectory.

The experiments follow the same conditional prediction perspective. A synthetic branching system tests recovery of both population distributions and branching probabilities. Comparisons across numerical sampling budgets assess the balance between one-step efficiency and prediction accuracy. Finally, mouse hematopoiesis and direct reprogramming demonstrate the use of lineage information for cell fate prediction, with improved probability scores relative to LIFT trained with geometric couplings.

We organize this paper as follows. Section~\ref{sec:problem_formulation} formulates the reconstruction problem, Section~\ref{sec:latent_dynamics} develops LIFT, and Section~\ref{sec:error_analysis} establishes its error estimates. Section~\ref{sec:experiments} presents the experiments, followed by Section~\ref{sec:discussion_conclusion} on conclusions and future work.

\section{Problem Formulation and Conditional Fate Reconstruction}
\label{sec:problem_formulation}

We consider trajectory reconstruction from barcode-resolved cell populations observed at two time points. The endpoint measures are given, but the correspondence between individual source and terminal states remains unknown. We distinguish the true lineage-compatible coupling, denoted by $\pi_\star$, from the coupling $\pi$ used for training. Our objective is to recover the conditional transition structure of $\pi_\star$.

\subsection{Barcode-resolved observations and the target dynamics}
\label{subsec:mathematical_formulation}

Consider two observation times, rescaled to $0$ and $1$, and
$M$ distinct barcodes. For each barcode $i=1, 2, \cdots,M$, let $\mu_0^i,\mu_1^i$ be the observed nonnegative finite measures on $\mathbb R^d$
describing its source and terminal cell populations. Assume that
these measures have finite second moments. The total population
measures are
\begin{equation}
\mu_0=\sum_{i=1}^M\mu_0^i,
\qquad
\mu_1=\sum_{i=1}^M\mu_1^i,
\label{eq:barcode_population_measures}
\end{equation}
where we assume $\mu_0^i(\mathbb{R}^d) =\mu_1^i(\mathbb{R}^d)=w_i, \sum w_i=1$, so that $\mu_0, \mu_1$ are probability measures, and $\pi_\star$ becomes a probability measure as well.

The true endpoint relationship is represented by an unknown coupling
$\pi_\star$ with marginals $\mu_0$ and $\mu_1$. Since barcode identity
is preserved, this coupling permits no transport between different
barcodes. More precisely, it admits a decomposition
\begin{equation}
\pi_\star=\sum_{i=1}^M\pi_\star^i,
\qquad
\pi_\star^i\in\Pi(\mu_0^i,\mu_1^i),
\label{eq:true_barcode_coupling}
\end{equation}
where $\Pi(\mu_0^i,\mu_1^i)$ denotes the finite nonnegative measures
on $\mathbb R^d\times\mathbb R^d$ with the indicated marginals.
Equation~\eqref{eq:true_barcode_coupling} is a block-diagonal
constraint with respect to barcode labels, as all blocks with distinct labels vanish.

Define the set of lineage compatible couplings by
\begin{equation}
\mathcal C_{\mathrm{lin}}
=
\left\{
\sum_{i=1}^M\gamma^i:
\gamma^i\in\Pi(\mu_0^i,\mu_1^i)
\right\}
\subseteq\Pi(\mu_0,\mu_1).
\label{eq:lineage_coupling_set}
\end{equation}
The observations specify this constraint set, but do not generally
identify its true member $\pi_\star$. In practice, training sampling uses a
constructed coupling $
\pi\in\Pi(\mu_0,\mu_1),
$ chosen from $\mathcal C_{\mathrm{lin}}$ when the barcode constraints are enforced.

Because the two couplings have the same marginals, we can write their conditional decompositions as follows.
\begin{equation}
\begin{aligned}
\pi_\star(\mathrm dx,\mathrm dy)
&=\mu_0(\mathrm dx)K_\star(x,\mathrm dy),\\
\pi(\mathrm dx,\mathrm dy)
&=\mu_0(\mathrm dx)K_\pi(x,\mathrm dy).
\end{aligned}
\label{eq:true_and_training_disintegration}
\end{equation}
Here, $K_\star(x,\cdot)$ is the true distribution of terminal states
conditioned on the observed molecular state $x$, whereas
$K_\pi(x,\cdot)$ is the corresponding conditional distribution
induced by the training coupling. For a measurable partition
$\{B_1,\ldots,B_K\}$ of the terminal state space into distinct fates,
the true fate probabilities are
\begin{equation}
p_k^\star(x)=K_\star(x,B_k).
\label{eq:true_fate_probability}
\end{equation}

To reconstruct continuous dynamics, we seek a probability law on $\Omega=C([0,1];\mathbb R^d)$ whose initial-terminal joint distribution is $\pi_\star$.
Writing $e_t(\gamma)=\gamma(t)$ for the evaluation map, the ideal
endpoint requirement is
\begin{equation}
(e_0,e_1)_\#\mathbb P=\pi_\star.
\label{eq:true_endpoint_path_constraint}
\end{equation}
The endpoint coupling does not uniquely determine intermediate paths;
their construction therefore requires additional dynamical
assumptions.

For probability measures $P,Q$ with finite second
moments, we can define the Wassernstein distance
\[
W_2^2(P,Q)
=
\inf_{\lambda\in\Pi(P,Q)}
\int\|a-b\|^2\,\lambda(\mathrm da,\mathrm db),
\]
where $\Pi(P,Q)$ contains all pairings with marginals $P$ and $Q$, to evaluate the distance between two measures. For a learned path law $\widehat{\mathbb P}$ with initial marginal
$\mu_0$, define
\[
\widehat K_{0,t}(x,A)
=
\widehat{\mathbb P}(e_t\in A\mid e_0=x),
\qquad
\widehat K_1=\widehat K_{0,1}.
\]
Assuming that the generated terminal distribution has a finite
second moment, we measure the true conditional reconstruction
error by
\begin{equation}
\mathcal E_\star^2(\widehat K_1)
=
\int W_2^2\!\left(
K_\star(x,\cdot),\widehat K_1(x,\cdot)
\right)\mu_0(\mathrm dx)
\label{eq:true_conditional_error}
\end{equation}
The Wasserstein triangle inequality and Minkowski's inequality yield
\[
\mathcal E_\star(\widehat K_1)
\le
\varepsilon_{\mathrm{cpl}}
+
\mathcal E_\pi(\widehat K_1).
\]
where 
\begin{equation}
\mathcal E_\pi^2(\widehat K_1)
=
\int W_2^2\!\left(
K_\pi(x,\cdot),\widehat K_1(x,\cdot)
\right)\mu_0(\mathrm dx),
\varepsilon_\mathrm{cpl}^2
=
\int W_2^2\!\left(
K_\star(x,\cdot), K_\pi(x,\cdot)
\right)\mu_0(\mathrm dx)
\label{eq:error_decomposition}
\end{equation}
represent two sources of $\mathcal E_{\star}$. 
This discrepancy compares future distributions while holding the initial state fixed. In particular, matching the terminal marginal $\mu_1$ does not ensure recovery of the true conditional relationship.

Before analyzing training objectives, we examine a representational limitation that persists even if the true coupling were available.

\subsection{Limitations of deterministic mappings}
\label{subsec:deterministic_limitations}

A deterministic state generator produces an endpoint $\widehat Y=T(X)$, so its terminal conditional law is $ \widehat K_1(x,\cdot)=\delta_{T(x)}.$ This includes deterministic ODE models with unique solutions when the initial condition is fully specified by the observed state $x$. Their ability to recover the target coupling is limited by the conditional variability under $\pi_\star$.

Let
\[
m_\star(x)=\mathbb E_{\pi_\star}[Y\mid X=x],
\qquad
\Sigma_\star(x)
=
\operatorname{Cov}_{\pi_\star}(Y\mid X=x)
\]
denote the true conditional mean and covariance.

\begin{proposition}[Irreducible conditional error of deterministic maps]
\label{prop:deterministic_conditional_error}
For every measurable $T\in L^2(\mu_0;\mathbb R^d)$,
\begin{equation}
\begin{aligned}
\mathcal E_\star^2(T)
&:=
\int
W_2^2\!\left(
K_\star(x,\cdot),\delta_{T(x)}
\right)\mu_0(\mathrm dx)\\
&=
\int\operatorname{tr}\Sigma_\star(x)\,\mu_0(\mathrm dx)
+
\int\|m_\star(x)-T(x)\|^2\,\mu_0(\mathrm dx).
\end{aligned}
\label{eq:deterministic_error_decomposition}
\end{equation}
Consequently,
\begin{equation}
\inf_{T\in L^2(\mu_0;\mathbb R^d)}
\mathcal E_\star^2(T)
=
\int\operatorname{tr}\Sigma_\star(x)\,\mu_0(\mathrm dx),
\label{eq:irreducible_deterministic_error}
\end{equation}
with the minimum attained by $T=m_\star$.

Moreover, an exact deterministic realization
\[
\pi_\star=(\operatorname{Id},T)_\#\mu_0
\]
exists if and only if $K_\star(x,\cdot)$ is a Dirac measure for
$\mu_0$-almost every $x$.
\end{proposition}

We provide proof of the proposition in \ref{app:prop_proof}. The proposition separates deterministic reconstruction error into the true conditional variability and the error in approximating the true conditional mean. Improving a deterministic model can reduce the second component, but not the first.

Crucially, this limitation is conditional rather than marginal. A deterministic flow can produce a terminal population by sending different initial states to different regions. However, it produces
only one endpoint from each fixed initial state. Its fate
probabilities under the prescribed partition are therefore
\[
\widehat p_k(x)=\mathbf 1_{B_k}(T(x)),
\]
which cannot equal a true probability
$0<p_k^\star(x)<1$ without additional randomization.

Motivated by these observations, we introduce a latent representation,
\[
Z\sim p_\eta(\cdot\mid x),
\qquad
\widehat Y=F_\theta(x,Z),
\]
whose terminal conditional law is
\[
\widehat K_1(x,\cdot)
=
F_\theta(x,\cdot)_\#p_\eta(\cdot\mid x).
\]
Unlike a deterministic state map, this representation allows
multiple future outcomes from the same observed state. The following sections develop this representation into continuous dynamics and relate its training error to the conditional reconstruction objective.

\section{Variational Learning of Latent Cell Dynamics}
\label{sec:latent_dynamics}

We now construct a probabilistic dynamical model from the training
coupling $\pi$. A latent variable represents alternative outcomes
from the same observed initial state, while a time-dependent decoder
generates the corresponding cell-state trajectories. We first derive
a variational reconstruction objective, then introduce flow matching
and Mean Flow as simulation-free training strategies. Throughout
this section, training expectations are taken under $\pi$, whereas
the true conditional reconstruction error remains defined relative
to $\pi_\star$ as in
\eqref{eq:true_conditional_error}.

\subsection{Variational formulation}
\label{subsec:variational_formulation}

Let $Z\in\mathbb R^{d_z}$ be a latent variable. The generative model
consists of a conditional prior $p_\eta(\mathrm dz\mid x)$ and a
decoder $F_\theta(x,z)$, giving
\begin{equation}
Z\sim p_\eta(\cdot\mid x),
\qquad
\widehat Y=F_\theta(x,Z).
\label{eq:latent_endpoint_generator}
\end{equation}
Its terminal conditional distribution is
\begin{equation}
\widehat K_1(x,\cdot)
=
F_\theta(x,\cdot)_\#p_\eta(\cdot\mid x).
\label{eq:latent_terminal_kernel}
\end{equation}
During training, an encoder $q_\phi(\mathrm dz\mid x,y)$ uses both
endpoints to infer latent representations compatible with the
observed pair. During inference, the terminal state is unavailable,
and latent variables are drawn from $p_\eta(\cdot\mid x)$.

We parameterize the encoder and conditional prior by diagonal
Gaussian distributions,
\begin{equation}
\begin{aligned}
q_\phi(\cdot\mid x,y)
&=
\mathcal N\!\left(
m_\phi(x,y),
\operatorname{diag}(s_\phi^2(x,y))
\right),\\
p_\eta(\cdot\mid x)
&=
\mathcal N\!\left(
m_\eta(x),
\operatorname{diag}(s_\eta^2(x))
\right),
\end{aligned}
\label{eq:gaussian_encoder_prior}
\end{equation}
with strictly positive standard deviations. The means and standard
deviations are outputs of neural networks. Define the training joint
distribution
\begin{equation}
Q_\phi(\mathrm dx,\mathrm dy,\mathrm dz)
=
\pi(\mathrm dx,\mathrm dy)
q_\phi(\mathrm dz\mid x,y).
\label{eq:training_augmented_distribution}
\end{equation}

To derive a variational objective, introduce the auxiliary
observation model
\begin{equation}
p_\theta^\tau(y\mid x,z)
=
\mathcal N\!\left(y;F_\theta(x,z),\tau^2I_d\right),
\qquad \tau>0,
\label{eq:auxiliary_gaussian_decoder}
\end{equation}
where $\tau$ is fixed. Its conditional likelihood is
\[
p_{\theta,\eta}^\tau(y\mid x)
=
\int
p_\theta^\tau(y\mid x,z)p_\eta(z\mid x)\,\mathrm dz.
\]
Following variational inference
\cite{Kingma2014VAE,Sohn2015CVAE}, Jensen's inequality gives
\begin{equation}
\begin{aligned}
\log p_{\theta,\eta}^\tau(y\mid x)
&=
\log\mathbb E_{q_\phi(\cdot\mid x,y)}
\left[
\frac{
p_\theta^\tau(y\mid x,Z)p_\eta(Z\mid x)
}{
q_\phi(Z\mid x,y)
}
\right]\\
&\ge
\mathbb E_{q_\phi(\cdot\mid x,y)}
\log p_\theta^\tau(y\mid x,Z)
-
D_{\mathrm{KL}}\!\left(
q_\phi(\cdot\mid x,y)\|
p_\eta(\cdot\mid x)
\right).
\end{aligned}
\label{eq:conditional_elbo}
\end{equation}

Define the reconstruction error and prior-matching penalty by
\begin{equation}
\begin{aligned}
D_\pi(\theta,\phi)
&=
\mathbb E_{Q_\phi}
\|Y-F_\theta(X,Z)\|^2,\\
\mathcal R_{\mathrm{KL}}(\phi,\eta)
&=
\mathbb E_{\pi}
D_{\mathrm{KL}}\!\left(
q_\phi(\cdot\mid X,Y)\|
p_\eta(\cdot\mid X)
\right).
\end{aligned}
\label{eq:reconstruction_and_kl}
\end{equation}
Substituting the Gaussian observation model into
\eqref{eq:conditional_elbo} and averaging over $\pi$ yields
\begin{equation}
-\mathbb E_\pi
\log p_{\theta,\eta}^\tau(Y\mid X)
\le
\frac d2\log(2\pi\tau^2)
+
\frac{D_\pi(\theta,\phi)}{2\tau^2}
+
\mathcal R_{\mathrm{KL}}(\phi,\eta).
\label{eq:negative_loglikelihood_bound}
\end{equation}
Dropping the parameter-independent constant gives the negative
evidence lower bound objective
\begin{equation}
\mathcal L_{\mathrm{VI}}
=
\frac{1}{2\tau^2}D_\pi
+
\mathcal R_{\mathrm{KL}}.
\label{eq:variational_objective}
\end{equation}

The Gaussian model in
\eqref{eq:auxiliary_gaussian_decoder} provides a likelihood-based
interpretation of the reconstruction objective. The generator
considered below remains \eqref{eq:latent_endpoint_generator},
without additional observation noise.

Direct evaluation of $D_\pi$ can be expensive when $F_\theta$ is defined through a differential equation. We therefore seek simulation-free objectives that control this reconstruction error without solving the learned dynamics during training.

\subsection{Flow matching and Mean Flow decoders}
\label{subsec:flow_decoders}

Both training strategies use endpoint pairs drawn from $\pi$ and
latent variables drawn from the encoder. For
$(X,Y,Z)\sim Q_\phi$, define the reference interpolation
\begin{equation}
X_t=(1-t)X+tY,
\qquad
V=Y-X.
\label{eq:training_reference_interpolation}
\end{equation}
The latent variable is held fixed along each reference path.
The interpolation specifies a training path between the endpoints, and is not assumed to be an observed biological trajectory.

\paragraph{Flow matching}

Flow matching trains a time-dependent velocity field by regression
against velocities of prescribed reference paths, avoiding
numerical integration during training
\cite{Lipman2023FM,Tong2024CFM}.
In our latent formulation, the decoder is defined by
\begin{equation}
\frac{\mathrm d}{\mathrm dt}
\varphi_{\theta,t}^z(x)
=
v_\theta(\varphi_{\theta,t}^z(x),z,t),
\qquad
\varphi_{\theta,0}^z(x)=x,
\label{eq:latent_flow_ode}
\end{equation}
with terminal map
\[
F_\theta^{\mathrm{FM}}(x,z)
=
\varphi_{\theta,1}^z(x).
\]
The conditional flow-matching objective is
\begin{equation}
\mathcal L_{\mathrm{FM}}(\theta,\phi)
=
\mathbb E_{\substack{
(X,Y,Z)\sim Q_\phi\\
t\sim U[0,1]
}}
\left\|
v_\theta(X_t,Z,t)-V
\right\|^2.
\label{eq:latent_fm_loss}
\end{equation}
Its inputs and regression targets are available directly from
\eqref{eq:training_reference_interpolation}. For a fixed encoder,
the unrestricted squared-loss minimizer is
\[
\overline v_\phi(\xi,z,t)
=
\mathbb E_{Q_\phi}[V\mid X_t=\xi,Z=z].
\]
Conditioning on $Z$ allows the regression to distinguish different
outcomes that may overlap in the observed state space.

The FM loss is evaluated along the prescribed reference paths, whereas generation follows the learned ODE. Under suitable regularity assumptions, the former controls the endpoint reconstruction error of the latter. See Theorem~\ref{thm:fm_reconstruction_control} in Section~\ref{sec:error_analysis}.

\paragraph{Mean Flow}

Mean Flow replaces instantaneous velocity with an interval-average
velocity, allowing direct prediction of finite-time displacements
\cite{Geng2025MeanFlow}. Using a forward-time convention, consider
a trajectory satisfying $\dot x_s=v(x_s,z,s)$ and define
\begin{equation}
u(x_r,z;r,t)
=
\frac{1}{t-r}\int_r^t v(x_s,z,s)\,\mathrm ds
=
\frac{x_t-x_r}{t-r},
\qquad r<t.
\label{eq:mean_velocity_definition}
\end{equation}
Consequently,
\[
x_t=x_r+(t-r)u(x_r,z;r,t).
\]
Differentiating along the trajectory with respect to the starting
time $r$ gives
\begin{equation}
u(x_r,z;r,t)
=
v(x_r,z,r)
+
(t-r)
\left[
\partial_r u
+
D_xu\,v(x_r,z,r)
\right],
\label{eq:forward_mean_flow_identity}
\end{equation}
where all derivatives on the right are evaluated at
$(x_r,z;r,t)$.

We parameterize the average velocity by $u_\theta$ and use the
reference velocity $V$ to construct the training target.
Define the directional derivative
\begin{equation}
\mathcal D_r u_\theta
=
\partial_r u_\theta(X_r,Z;r,t)
+
D_xu_\theta(X_r,Z;r,t)V
\label{eq:mf_directional_derivative}
\end{equation}
and the target
\[
\mathcal T_\theta
=
V+(t-r)\mathcal D_r u_\theta.
\]
The Mean Flow loss is
\begin{equation}
\mathcal L_{\mathrm{MF}}(\theta,\phi)
=
\mathbb E_{\substack{
(X,Y,Z)\sim Q_\phi\\
t\sim U[0,1],\ r\mid t\sim U[0,t]
}}
\left\|
u_\theta(X_r,Z;r,t)
-
\operatorname{sg}(\mathcal T_\theta)
\right\|^2,
\label{eq:latent_mf_loss}
\end{equation}
where $\operatorname{sg}$ denotes the stop-gradient operation.
The directional derivative is computed by a Jacobian vector
product, without constructing a full Jacobian. Detaching the
target avoids differentiating through this derivative when
updating the model.

The numerical value of
\eqref{eq:latent_mf_loss} is the squared residual of the
reference-path identity:
\[
\mathcal L_{\mathrm{MF}}
=
\mathbb E
\left\|
u_\theta(X_r,Z;r,t)
-
V-(t-r)\mathcal D_r u_\theta
\right\|^2.
\]
This observation allows its value to be related to reconstruction
error independently of the stop-gradient update rule.

The learned conditional path and endpoint map are
\begin{equation}
G_{\theta,t}(x,z)
=
x+t\,u_\theta(x,z;0,t),
\qquad
F_\theta^{\mathrm{MF}}(x,z)
=
G_{\theta,1}(x,z).
\label{eq:mf_direct_decoder}
\end{equation}

We use the Mean Flow loss as a simulation-free surrogate for the variational reconstruction term. Its control of the one-step endpoint error is established in Theorem~\ref{thm:mf_reconstruction_control}. Thus, the resulting training objective is motivated by the variational formulation, but is not algebraically identical to the negative evidence lower bound.

\subsection{Training and sampling}
\label{subsec:training_sampling}

We use the Mean Flow decoder for direct conditional generation.
In addition to the prior-matching penalty, we regularize the encoder
against a standard Gaussian:
\begin{equation}
\mathcal R_0(\phi)
=
\mathbb E_\pi
D_{\mathrm{KL}}\!\left(
q_\phi(\cdot\mid X,Y)\|
\mathcal N(0,I_{d_z})
\right).
\label{eq:latent_reference_regularizer}
\end{equation}

The full training objective is
\begin{equation}
\mathcal L_{\mathrm{LIFT}}(\theta,\phi,\eta)
=
C\,\mathcal L_{\mathrm{MF}}(\theta,\phi)
+
\mathcal R_{\mathrm{KL}}(\phi,\eta)
+
\beta\,\mathcal R_0(\phi),
\label{eq:lift_training_objective}
\end{equation}
where $C>0$ and $\beta\ge0$. The KL terms are evaluated analytically
for the Gaussian parameterization, while the Mean Flow term is
estimated from sampled endpoint pairs, latent variables, and times.

Latent samples are obtained by reparameterization:
\begin{equation}
Z
=
m_\phi(X,Y)
+
s_\phi(X,Y)\odot\epsilon,
\qquad
\epsilon\sim\mathcal N(0,I_{d_z}),
\label{eq:encoder_reparameterization}
\end{equation}
which permits joint updates of the encoder, prior, and decoder.
Gradients through $Z$ are retained in the direct decoder evaluation;
the Mean Flow regression target is detached as in
\eqref{eq:latent_mf_loss}.

Endpoint pairs are sampled according to the prescribed coupling $\pi$. When $\pi=\sum_i\pi^i$ is lineage-compatible, this can be implemented by sampling barcode $i$ with probability $w_i$ and then sampling a pair from $\pi^i/w_i$, for $w_i>0$. Thus, barcode information enters through the training coupling without requiring barcode labels as network inputs.

\begin{algorithm}[t]
\caption{LIFT training}
\label{alg:lift_training}
\begin{algorithmic}[1]
\REQUIRE Training coupling $\pi$, batch size $B$, weights $C,\beta$.
\STATE Initialize encoder $\phi$, conditional prior $\eta$,
and Mean Flow decoder $\theta$.
\WHILE{the training budget is not exhausted}
    \STATE Sample $\{(x_b,y_b)\}_{b=1}^B$ independently from $\pi$.
    \FOR{$b=1,\ldots,B$}
        \STATE Evaluate $q_b=q_\phi(\cdot\mid x_b,y_b)$
        and $p_b=p_\eta(\cdot\mid x_b)$.
        \STATE Sample $\epsilon_b\sim\mathcal N(0,I_{d_z})$
        and set $z_b=m_\phi(x_b,y_b)
        +s_\phi(x_b,y_b)\odot\epsilon_b$.
        \STATE Sample $t_b\sim U[0,1]$
        and $r_b\sim U[0,t_b]$.
        \STATE Set $x_{r_b}=(1-r_b)x_b+r_by_b$
        and $V_b=y_b-x_b$.
        \STATE Evaluate $U_b=u_\theta(x_{r_b},z_b;r_b,t_b)$.
        \STATE Compute
        $J_b=\partial_r u_\theta+
        D_xu_\theta V_b$
        at $(x_{r_b},z_b;r_b,t_b)$
        using a Jacobian--vector product.
        \STATE Set
        $T_b=\operatorname{sg}\!\left(
        V_b+(t_b-r_b)J_b\right)$.
        \STATE Compute
        $\ell_b=
        C\|U_b-T_b\|^2+
        D_{\mathrm{KL}}(q_b\|p_b)+
        \beta D_{\mathrm{KL}}(q_b\|\mathcal N(0,I_{d_z}))$.
    \ENDFOR
    \STATE Update $(\theta,\phi,\eta)$ using
    $B^{-1}\sum_{b=1}^B\ell_b$ and the specified
    stop-gradient rule.
\ENDWHILE
\RETURN Trained parameters $(\theta,\phi,\eta)$.
\end{algorithmic}
\end{algorithm}

At inference, only the conditional prior and Mean Flow decoder are
required. Given an initial state $x$, draw
$Z\sim p_\eta(\cdot\mid x)$ and evaluate
\begin{equation}
\widehat X_t
=
x+t\,u_\theta(x,Z;0,t).
\label{eq:lift_sampling_path}
\end{equation}
The same latent sample is used at all queried times to represent
one realization of a path. A different latent sample produces
another possible realization from the same initial state.

If $u_\theta$ is continuous in time, the map $t\mapsto\widehat X_t$ is continuous and satisfies $\widehat X_0=x$. This construction defines a conditional path law. It does not require an independently trained instantaneous velocity field or numerical ODE integration.

\begin{algorithm}[t]
\caption{LIFT conditional sampling}
\label{alg:lift_sampling}
\begin{algorithmic}[1]
\REQUIRE Initial state $x$, query times
$\{t_\ell\}_{\ell=1}^L\subseteq[0,1]$,
number of realizations $N$, trained parameters $(\theta,\eta)$.
\STATE Evaluate $m=m_\eta(x)$ and $s=s_\eta(x)$.
\FOR{$n=1,\ldots,N$}
    \STATE Sample $\epsilon_n\sim\mathcal N(0,I_{d_z})$
    and set $z_n=m+s\odot\epsilon_n$.
    \FOR{$\ell=1,\ldots,L$}
        \STATE Set
        $\widehat X_{t_\ell}^{(n)}
        =
        x+t_\ell u_\theta(x,z_n;0,t_\ell)$.
    \ENDFOR
\ENDFOR
\RETURN Sampled states
$\{\widehat X_{t_\ell}^{(n)}\}_{n,\ell}$.
\end{algorithmic}
\end{algorithm}

After evaluating the conditional prior, sampling requires one
decoder evaluation per realization and requested time point.
For a terminal fate region $B_k$, its model probability can be
estimated by
\begin{equation}
\widehat p_k^{(N)}(x)
=
\frac1N\sum_{n=1}^N
\mathbf 1_{B_k}\!\left(\widehat X_1^{(n)}\right).
\label{eq:mc_fate_estimator}
\end{equation}
This estimates
$\widehat K_1(x,B_k)$; increasing $N$ reduces Monte Carlo error
but does not remove coupling or model approximation errors.

The resulting model is trained using the available coupling $\pi$ and sampled through the learned conditional prior. Section~\ref{sec:error_analysis} relates its training losses to reconstruction under the unknown coupling $\pi_\star$ and to the accuracy of the resulting fate-probability estimates.

\section{Error Analysis of Conditional Dynamics and Fate Prediction}
\label{sec:error_analysis}

We analyze how the training objectives control conditional reconstruction and fate prediction. The analysis proceeds in three steps. First, the FM and MF losses control reconstruction under the training coupling $\pi$. Second, reconstruction and latent distribution matching are related to the true conditional error under $\pi_\star$. Finally, this error is translated into fate-probability error and combined with the uncertainty of finite Monte Carlo sampling.

Throughout this section, the model parameters are fixed, and all
training losses denote expectations under the prescribed training
distribution rather than individual minibatch estimates. The
couplings $\pi$ and $\pi_\star$ have the same marginals $\mu_0$
and $\mu_1$. We retain the notation
\[
Q_\phi(\mathrm dx,\mathrm dy,\mathrm dz)
=
\pi(\mathrm dx,\mathrm dy)
q_\phi(\mathrm dz\mid x,y)
\]
from Section~\ref{sec:latent_dynamics}.
All proofs of the results in this section are provided in \ref{app:error_proofs}.

\subsection{Reconstruction guarantees for flow matching and Mean Flow}
\label{subsec:reconstruction_error}

For either decoder, the training reconstruction error is
\begin{equation}
D_\pi
=
\mathbb E_{Q_\phi}
\|Y-F_\theta(X,Z)\|^2.
\label{eq:analysis_reconstruction_error}
\end{equation}
Recall that the training reference path is
\[
X_t=(1-t)X+tY,
\qquad
V=Y-X.
\]
The following results connect the regression losses evaluated along
this reference path to the endpoint produced by the learned decoder.

\paragraph{Flow matching}

For the FM decoder, let
\[
\frac{\mathrm d}{\mathrm dt}\widehat X_t
=
v_\theta(\widehat X_t,Z,t),
\qquad
\widehat X_0=X,
\]
and set $F_\theta^{\mathrm{FM}}(X,Z)=\widehat X_1$.
Its training loss is
\[
\mathcal L_{\mathrm{FM}}
=
\mathbb E_{Q_\phi}
\int_0^1
\|v_\theta(X_t,Z,t)-V\|^2\,\mathrm dt.
\]
Although this loss is evaluated on the reference interpolation,
the reconstruction error concerns the solution of the learned ODE.
A stability estimate bridges this difference.

\begin{theorem}[Reconstruction control by flow matching]
\label{thm:fm_reconstruction_control}
Assume that $v_\theta$ is measurable in time and globally
$L_x$-Lipschitz in its state argument, uniformly in $z$ and $t$.
Assume also that $t\mapsto v_\theta(0,z,t)$ is integrable on
$[0,1]$ for almost every latent variable under the training law,
so that the generated ODE has a unique solution.
If $\mathcal L_{\mathrm{FM}}<\infty$, then
\begin{equation}
D_\pi^{\mathrm{FM}}
:=
\mathbb E_{Q_\phi}
\|Y-F_\theta^{\mathrm{FM}}(X,Z)\|^2
\le
C_{L_x}\mathcal L_{\mathrm{FM}},
\label{eq:fm_reconstruction_bound}
\end{equation}
where
\begin{equation}
C_{L_x}
=
\begin{cases}
\dfrac{e^{2L_x}-1}{2L_x},&L_x>0,\\[2mm]
1,&L_x=0.
\end{cases}
\label{eq:fm_stability_constant}
\end{equation}
\end{theorem}

The estimate accounts for the accumulation of local regression errors along the generated trajectory. It concerns the exact ODE solution; a numerical implementation introduces an additional integration error.

\paragraph{Mean Flow}

For the Mean Flow decoder,
\[
F_\theta^{\mathrm{MF}}(x,z)
=
x+u_\theta(x,z;0,1).
\]
Define the residual along the reference path by
\begin{equation}
\begin{aligned}
\mathscr R_{r,t}
={}&
u_\theta(X_r,Z;r,t)-V\\
&-(t-r)\left[
\partial_r u_\theta(X_r,Z;r,t)
+
D_xu_\theta(X_r,Z;r,t)V
\right].
\end{aligned}
\label{eq:mf_reference_residual}
\end{equation}
The numerical value of the Mean Flow loss is
\begin{equation}
\mathcal L_{\mathrm{MF}}
=
\mathbb E_{\substack{
(X,Y,Z)\sim Q_\phi\\
t\sim U[0,1],\ r\mid t\sim U[0,t]
}}
\|\mathscr R_{r,t}\|^2.
\label{eq:mf_residual_loss}
\end{equation}
The stop-gradient operation used in training changes the parameter
update but not this residual value.

\begin{theorem}[Reconstruction control by Mean Flow]
\label{thm:mf_reconstruction_control}
Assume that, for $Q_\phi$-almost every $(X,Y,Z)$ and every
$t\in(0,1]$, the map
\[
r\longmapsto u_\theta(X_r,Z;r,t)
\]
is absolutely continuous on $[0,t]$, with derivative
\[
\partial_r u_\theta(X_r,Z;r,t)
+
D_xu_\theta(X_r,Z;r,t)V
\]
almost everywhere.

Assume further that
$t\mapsto u_\theta(X,Z;0,t)$ is Lipschitz as a map into
$L^2(Q_\phi;\mathbb R^d)$:
\begin{equation}
\left\|
u_\theta(X,Z;0,t)
-
u_\theta(X,Z;0,s)
\right\|_{L^2(Q_\phi)}
\le
L_T|t-s|,
\qquad s,t\in[0,1].
\label{eq:mf_time_regularity}
\end{equation}
If $\mathcal L_{\mathrm{MF}}<\infty$, then
\begin{equation}
D_\pi^{\mathrm{MF}}
:=
\mathbb E_{Q_\phi}
\|Y-F_\theta^{\mathrm{MF}}(X,Z)\|^2
\le
\max\left\{
3\mathcal L_{\mathrm{MF}},
(3L_T\mathcal L_{\mathrm{MF}})^{2/3}
\right\}.
\label{eq:mf_reconstruction_bound}
\end{equation}
\end{theorem}

For a unified notation, define
\begin{equation}
\begin{aligned}
\Psi_{\mathrm{FM}}(a)
=C_{L_x}a,\quad
\Psi_{\mathrm{MF}}(a)
=\max\{3a,(3L_Ta)^{2/3}\}.
\end{aligned}
\label{eq:decoder_error_moduli}
\end{equation}
Then, for either decoder
$\mathrm{dec}\in\{\mathrm{FM},\mathrm{MF}\}$,
\begin{equation}
D_\pi^{\mathrm{dec}}
\le
\Psi_{\mathrm{dec}}(\mathcal L_{\mathrm{dec}}).
\label{eq:unified_reconstruction_bound}
\end{equation}

These estimates control endpoint reconstruction under $\pi$. They do not by themselves establish fidelity to the unknown coupling $\pi_\star$, nor do they identify the true intermediate biological paths. The next result addresses the former question.

\subsection{Total conditional reconstruction error}
\label{subsec:total_conditional_error}

There are two additional discrepancies between training reconstruction and the desired conditional generation. First, the training coupling $\pi$ may differ from $\pi_\star$. Second, training uses the encoder $q_\phi(\cdot\mid x,y)$, whereas sampling uses the conditional prior $p_\eta(\cdot\mid x)$.

Recall the conditional decompositions
\[
\pi_\star(\mathrm dx,\mathrm dy)
=
\mu_0(\mathrm dx)K_\star(x,\mathrm dy),
\qquad
\pi(\mathrm dx,\mathrm dy)
=
\mu_0(\mathrm dx)K_\pi(x,\mathrm dy),
\]
and write the conditional coupling discrepancy as
\begin{equation}
\varepsilon_{\mathrm{cpl}}^2
=
\int
W_2^2\!\left(
K_\star(x,\cdot),K_\pi(x,\cdot)
\right)\mu_0(\mathrm dx).
\label{eq:conditional_coupling_discrepancy}
\end{equation}

To describe the training latent distribution conditioned only on
the initial state, define the aggregated posterior
\begin{equation}
\overline q_\phi(\mathrm dz\mid x)
=
\int
q_\phi(\mathrm dz\mid x,y)K_\pi(x,\mathrm dy).
\label{eq:aggregated_conditional_posterior}
\end{equation}
Its discrepancy from the sampling prior is
\begin{equation}
\Delta_Z^2
=
\int
W_2^2\!\left(
\overline q_\phi(\cdot\mid x),
p_\eta(\cdot\mid x)
\right)\mu_0(\mathrm dx).
\label{eq:latent_sampling_discrepancy}
\end{equation}

The generated terminal kernel is
\[
\widehat K_1(x,\cdot)
=
F_\theta(x,\cdot)_\#p_\eta(\cdot\mid x).
\]
Its true conditional error is
\[
\mathcal E_\star^2(\widehat K_1)
=
\int
W_2^2\!\left(
K_\star(x,\cdot),\widehat K_1(x,\cdot)
\right)\mu_0(\mathrm dx).
\]

\begin{theorem}[Total conditional reconstruction error]
\label{thm:latent_true_conditional_error}
Assume that
\[
\|F_\theta(x,z)-F_\theta(x,z')\|
\le L_Z\|z-z'\|
\]
for $\mu_0$-almost every $x$ and all $z,z'$.
Assume that the required second moments and the quantities below
are finite. Then
\begin{equation}
\mathcal E_\star(\widehat K_1)
\le
\varepsilon_{\mathrm{cpl}}
+
\sqrt{D_\pi}
+
L_Z\Delta_Z.
\label{eq:latent_true_error_bound}
\end{equation}

Suppose additionally that the encoder and prior are diagonal
Gaussians as in \eqref{eq:gaussian_encoder_prior}, and that their
prior standard deviations satisfy
\[
0<s_{\eta,j}(x)\le S
\]
for every latent coordinate $j$ and $\mu_0$-almost every $x$.
Then
\begin{equation}
\Delta_Z^2
\le
2S^2\mathcal R_{\mathrm{KL}},
\label{eq:gaussian_latent_mismatch_bound}
\end{equation}
where
\[
\mathcal R_{\mathrm{KL}}
=
\mathbb E_\pi
D_{\mathrm{KL}}\!\left(
q_\phi(\cdot\mid X,Y)\|
p_\eta(\cdot\mid X)
\right).
\]
Consequently, under the assumptions of the corresponding
reconstruction theorem,
\begin{equation}
\mathcal E_\star(\widehat K_1)
\le
\varepsilon_{\mathrm{cpl}}
+
\sqrt{
\Psi_{\mathrm{dec}}(\mathcal L_{\mathrm{dec}})}
+
\sqrt{2}\,L_ZS
\sqrt{\mathcal R_{\mathrm{KL}}}.
\label{eq:training_to_true_error_bound}
\end{equation}
\end{theorem}

The three terms in \eqref{eq:latent_true_error_bound} have distinct interpretations. The coupling discrepancy measures the difference between the training relationships and the unknown target. The reconstruction term measures how accurately the decoder reproduces training endpoints from encoder samples. The final term measures the effect of replacing the training latent distribution with the prior used for generation.

These are conditional error guarantees for a fitted model. For later use, denote the upper bound in
\eqref{eq:training_to_true_error_bound} by
\begin{equation}
\mathcal B_{\mathrm{dec}}
=
\varepsilon_{\mathrm{cpl}}
+
\sqrt{
\Psi_{\mathrm{dec}}(\mathcal L_{\mathrm{dec}})
}
+
\sqrt{2}\,L_ZS\sqrt{\mathcal R_{\mathrm{KL}}}.
\label{eq:total_error_budget}
\end{equation}

\subsection{Fate-probability estimation}
\label{subsec:fate_probability_error}

Let $\{B_1,\ldots,B_K\}$ be the fixed partition of terminal
state space into fates. The true and model fate probabilities are
\begin{equation}
p_k^\star(x)=K_\star(x,B_k),
\qquad
\widehat p_k(x)=\widehat K_1(x,B_k).
\label{eq:true_model_fate_probabilities}
\end{equation}
Estimating $p_k^\star(x)$ by generated samples involves two
different errors: the discrepancy between $\widehat p_k(x)$ and
$p_k^\star(x)$, and the finite-sampling error in estimating
$\widehat p_k(x)$.

\paragraph{From conditional distributions to fate probabilities}

A small displacement can change a hard fate label when the terminal
state lies near a classification boundary. Define
\[
\Gamma=\bigcup_{k=1}^K\partial B_k,
\qquad
\Gamma_\rho
=
\{y:\operatorname{dist}(y,\Gamma)\le\rho\},
\qquad \rho>0.
\]
The next bound separates this boundary effect from the endpoint
reconstruction error.

\begin{proposition}[Stability of fate probabilities]
\label{prop:fate_probability_stability}
For every $\rho>0$,
\begin{equation}
\begin{aligned}
\max_{1\le k\le K}
\|p_k^\star-\widehat p_k\|_{L^1(\mu_0)}
&\le
\mu_1(\Gamma_\rho)
+
\frac{
\mathcal E_\star^2(\widehat K_1)
}{\rho^2}\\
&\le
\mu_1(\Gamma_\rho)
+
\frac{\mathcal B_{\mathrm{dec}}^2}{\rho^2},
\end{aligned}
\label{eq:hard_fate_probability_bound}
\end{equation}
where the second inequality holds under the assumptions of
Theorem~\ref{thm:latent_true_conditional_error}.
\end{proposition}

The estimate follows the geometry of a label change: a coupled
pair of endpoints can have different fates only if the true
endpoint lies near a boundary or the displacement between the
endpoints is sufficiently large. The radius $\rho$ balances
these two possibilities. In particular, if $\mu_1(\Gamma)=0$,
vanishing conditional reconstruction error implies vanishing
average fate-probability error.
Furthermore, we combine the results above with standard results of Monte-Carlo sampling\cite{Hoeffding1963}, we can obtain the following end-to-end fate probability error estimation.  
\begin{corollary}[End-to-end fate-probability error]
\label{cor:end_to_end_fate_error}
Suppose
\[
\mu_0=\sum_{a=1}^n\omega_a\delta_{x_a},
\qquad
\omega_a>0,
\qquad
\sum_{a=1}^n\omega_a=1.
\]
Under the assumptions of
Theorem~\ref{thm:latent_true_conditional_error},
with probability at least $1-\delta$ over the Monte Carlo samples,
\begin{equation}
\max_{1\le k\le K}
\sum_{a=1}^n\omega_a
\left|
\widehat p_k^{(N)}(x_a)-p_k^\star(x_a)
\right|\le
\inf_{\rho>0}
\left\{
\mu_1(\Gamma_\rho)
+
\frac{\mathcal B_{\mathrm{dec}}^2}{\rho^2}
\right\}
+
\sqrt{\frac{\log(2nK/\delta)}{2N}}.
\label{eq:end_to_end_fate_bound}
\end{equation}
\end{corollary}

\section{Experimental Validation and Lineage-Informed Applications}
\label{sec:experiments}
We evaluate LIFT on synthetic and real-world datasets. We first use a
synthetic branching benchmark to assess its ability to reconstruct
population distributions and branching probabilities. We then examine
whether one-step generation combines computational efficiency with
accurate conditional prediction. Finally, we evaluate how LIFT
incorporates lineage information to predict cell fate probabilities
in two real datasets. Across all experiments, we compare LIFT with baselines including conditional flow matching (CFM), Mean Flow (MF) and score matching (SM, here we use an implementation of SF2M\cite{tong2023simulationfree}) to analyze the ability of LIFT. Detailed experimental settings,
model-component ablations, hyperparameter sensitivity analyses, and
additional results are provided in Supplementary
Section~\ref{sec:sm_experiments}.

\subsection{Recovering conditional fate distributions in a synthetic dataset}
\label{subsec:synthetic_branching}

\paragraph{Dataset construction}
We sample 1,000 source states $x_i\sim\mathcal N(0,I_2)$ and assign
a branch by
\begin{equation}
S_i^j=\operatorname{sign}(v^\top x_i+\alpha\epsilon_i^j),
\qquad v=(1,2)^\top,\qquad\epsilon_i^j\sim\mathcal N(0,1),\qquad j=1, 2, \cdots,5.
\label{eq:synthetic_branch_rule}
\end{equation}
For each source state $x_i$, five targets $y_i^j$ are generated conditional on its
sampled branch $S_i^j$. Specifically, we construct $y_i^j = x_i + S_i^j \cdot 3v + 0.3\cdot\eta_i^j, \eta_i^j\sim \mathcal N(0,I_2)$. Thus, these five targets share the same barcode label, but may have different fate labels.  
For $\alpha>0$, the population probability of the positive branch is
\begin{equation}
p_\alpha^{\mathrm{br}}(x)
=\Phi\!\left(\frac{v^\top x}{\alpha}\right),
\label{eq:analytic_branch_probability}
\end{equation}
where $\Phi$ is the standard Gaussian distribution function. This analytic reference permits evaluation beyond the particular branch labels realized in the finite training sample.

The principal experiment uses $\alpha\in\{0.5,1,2\}$.
Branch probabilities are evaluated at $n_q$ uniformly spaced query states along the segment from $(-1,-1)$ to $(1,1)$. We define the probability error convention,
\begin{equation}
\mathrm{PE}
=\sum_{a=1}^{n_q}
\left|
\widehat p_+(x_a)-p_\alpha^{\mathrm{br}}(x_a)
\right|.
\label{eq:synthetic_probability_error}
\end{equation}
Here we choose $n_q=21$. Population level accuracy is measured by $W_2$ between generated and reference terminal samples.

\begin{table}[t]
\centering
\caption{Results on the synthetic branching benchmark.
Lower values are better. PE is the sum in
\eqref{eq:synthetic_probability_error}.}
\label{tab:synthetic_results}
\small
\begin{tabular}{lcccccc}
\toprule
&\multicolumn{2}{c}{$\alpha=0.5$}
&\multicolumn{2}{c}{$\alpha=1$}
&\multicolumn{2}{c}{$\alpha=2$}\\
\cmidrule(lr){2-3}\cmidrule(lr){4-5}\cmidrule(lr){6-7}
Method &$W_2$&PE&$W_2$&PE&$W_2$&PE\\
\midrule
CFM &\textbf{0.238}&1.370&0.275&2.678&0.305&5.002\\
Mean Flow &0.470&1.370&1.011&2.678&1.101&5.002\\
SM &0.463&0.282&0.478&1.500&0.460&3.580\\
LIFT &0.266&\textbf{0.121}&\textbf{0.268}&\textbf{0.457}
&\textbf{0.280}&\textbf{0.809}\\
\bottomrule
\end{tabular}
\end{table}

\begin{figure}[t]
\centering
    \begin{subfigure}{0.3\linewidth}
        \centering
        \includegraphics[width=\linewidth]{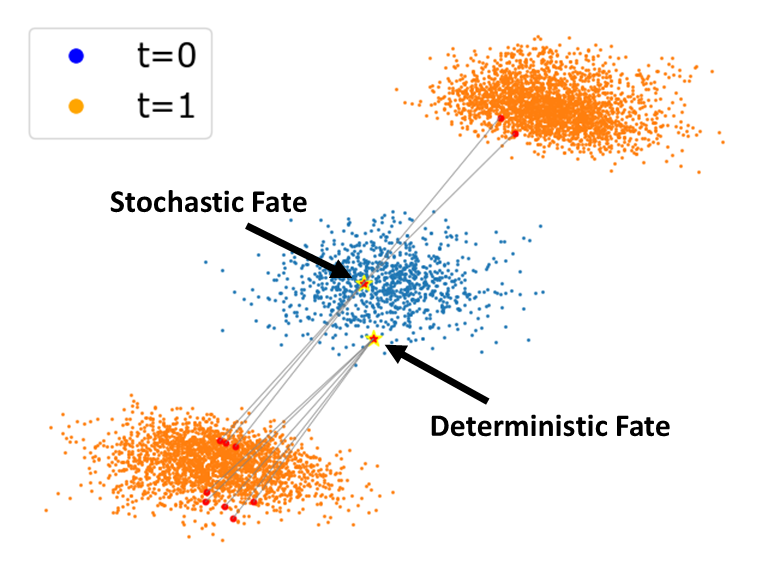}
        \caption{Dataset}
        \label{toy:sub1}
    \end{subfigure}
    \hfill 
    \begin{subfigure}{0.32\linewidth}
        \centering
        \includegraphics[width=\linewidth]{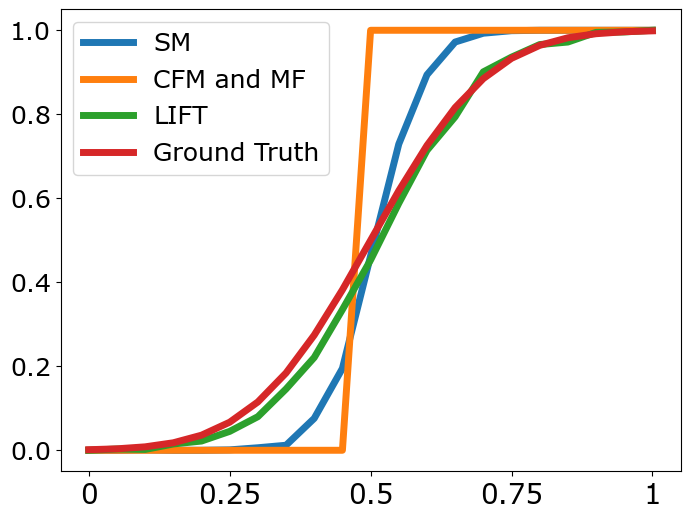}
        \caption{Predicted Upper Probability}
        \label{toy:sub2}
    \end{subfigure}
    \hfill 
    \centering
    \begin{subfigure}{0.34\linewidth}
        \centering
        \includegraphics[width=\linewidth]{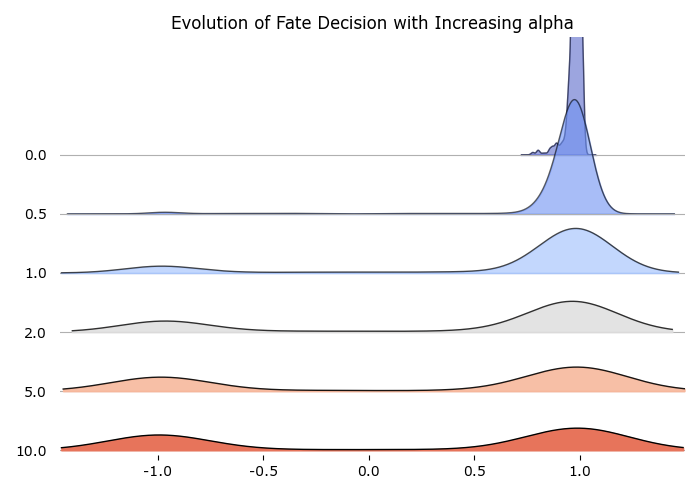}
        \caption{Conditional endpoint distributions}
        \label{toy:sub3}
    \end{subfigure}
    \hfill 

\caption{Conditional reconstruction in the branching system.
(a) Source and terminal populations. (b) Predicted branch probabilities
along the query segment. (c) Generated endpoint distributions at a
fixed initial state as $\alpha$ varies.}
\label{fig:synthetic_conditional}
\end{figure}

\paragraph{LIFT recovers both distribution and branching probabilities}
Table~\ref{tab:synthetic_results} shows that LIFT can achieve the lowest
branching probability error across all tested noise levels while
maintaining competitive population reconstruction. In contrast,
accurate marginal reconstruction by CFM does not translate into
accurate branching probabilities, illustrating the distinction
between these two objectives. Note that the CFM and MF probability errors coincide with the minimum error attainable by hard branch assignments on the evaluation grid, because deterministic methods can only provide 0-1 predictions. Figure~\ref{fig:synthetic_conditional}(b)
further shows that LIFT successfully captures the conditional branching probabilities, whereas the deterministic baselines
produce hard branch assignments.

\paragraph{Conditional distributions at a fixed initial state}
Figure~\ref{fig:synthetic_conditional}(c) examines the generated
endpoint distributions with the initial state held fixed.
At low noise, the predicted outcomes are concentrated in one branch. As $\alpha$ increases, probability mass is
distributed across both branches, producing a bimodal conditional
distribution. This qualitative transition
illustrates that LIFT can capture changes in branching uncertainty.

\subsection{Efficient one-step conditional generation}
\label{subsec:generation_efficiency}

We assess the accuracy-cost trade-off on the synthetic branching benchmark by varying the number of function evaluations (NFE) for CFM and SM. We compare these iterative configurations with one-step MF and LIFT, evaluating population reconstruction, branching probability error, and inference time at several noise levels.

\begin{table}[htbp]
\centering
\caption{Effect of NFE on the synthetic branching benchmark.
Inference times are averaged over 100 timing measurements.
Lower values are better for $W_2$, PE, and inference time.
The best values in each metric column are highlighted in bold.}
\label{tab:nfe_comparison}
\small
\setlength{\tabcolsep}{4pt}
\begin{tabular}{lrrcccccc}
\toprule
& & & \multicolumn{2}{c}{$\alpha=0.5$}
    & \multicolumn{2}{c}{$\alpha=1$}
    & \multicolumn{2}{c}{$\alpha=2$}\\
\cmidrule(lr){4-5}
\cmidrule(lr){6-7}
\cmidrule(lr){8-9}
Method & NFE & Time (ms)
& $W_2$ & PE & $W_2$ & PE & $W_2$ & PE\\
\midrule
CFM
& 1,000 & 401.09
& \textbf{0.238} & 1.370
& 0.275 & 2.678
& 0.305 & 5.002\\
& 100 & 40.26
& \textbf{0.238} & 1.370
& 0.282 & 2.678
& 0.305 & 5.002\\
& 10 & 3.87
& 0.270 & 1.370
& 0.367 & 2.678
& 0.369 & 5.002\\
\midrule
SM
& 1,000 & 855.93
& 0.463 & 0.282
& 0.478 & 1.500
& 0.460 & 3.580\\
& 100 & 86.74
& 0.531 & 0.382
& 0.548 & 1.556
& 0.558 & 3.619\\
& 10 & 9.65
& 0.628 & 0.623
& 0.656 & 1.501
& 0.719 & 3.834\\
\midrule
MF
& 1 & \textbf{0.56}
& 0.470 & 1.370
& 1.011 & 2.678
& 1.101 & 5.002\\
LIFT
& 1 & 0.96
& 0.266 & \textbf{0.121}
& \textbf{0.268} & \textbf{0.457}
& \textbf{0.280} & \textbf{0.809}\\
\bottomrule
\end{tabular}
\end{table}

\paragraph{LIFT balances generation efficiency and conditional accuracy}
Table~\ref{tab:nfe_comparison} shows that LIFT achieves the lowest
branching probability error across all tested noise levels while
sampling faster than every iterative configuration considered.
Its population reconstruction remains competitive, although CFM
with larger integration budgets attains a smaller $W_2$ in the
lowest-noise setting. Increasing NFE improves the marginal
reconstruction of the iterative baselines but does not consistently
improve their branching probabilities; in particular, the reported
CFM probability error remains high as it can give only 0-1 prediction. Although unaugmented MF
is slightly faster than LIFT, it has larger errors in both metrics.
These results show that LIFT combines efficient one-step sampling
with accurate branching probabilities. Specifically, lower inference time makes it convenient to predict fate probabilities through large-scale Monte-Carlo sampling.

\subsection{Lineage-informed cell fate prediction in real datasets}
\label{subsec:real_fate_prediction}

We evaluate LIFT on two lineage-traced scRNA-seq datasets: mouse hematopoiesis measured by LARRY \cite{Weinreb2020} and direct fibroblast reprogramming measured by CellTagging \cite{Biddy2018}. The hematopoietic task predicts monocyte (Mo) and neutrophil (Neu) outcomes from Day 2 to Day 6, while the reprogramming task predicts induced endoderm progenitor (iEP) and dead-end outcomes from Day 15 to Day 28. Both datasets use two-dimensional SPRING state representations. We compare geometric and lineage-informed couplings, denoted by Geo and Lin, respectively. The lineage-informed construction uses uniform pairing within each barcode for hematopoiesis and within-clone OT for reprogramming. Population reconstruction is measured by $W_2$, while fate prediction is assessed by mean probability error (MeanPE) and Pearson correlation $r$ against lineage-derived reference probabilities.

\begin{table}[htbp]
\centering
\caption{Results on the two lineage-traced datasets. Geo and Lin
indicate geometric and lineage-informed coupling variants,
respectively. Here SM relies on entropic optimal transport coupling as \cite{tong2023simulationfree}, so SM does not have different couplings. Lower $W_2$ and MeanPE and higher Pearson correlation
$r$ are better. The best value in each column is highlighted in bold.}
\label{tab:real_existing_results}
\small
\setlength{\tabcolsep}{4pt}
\begin{tabular}{lcccccc}
\toprule
& \multicolumn{3}{c}{Hematopoiesis}
& \multicolumn{3}{c}{Reprogramming}\\
\cmidrule(lr){2-4}
\cmidrule(lr){5-7}
Method & $W_2$ & MeanPE & $r$ & $W_2$ & MeanPE & $r$\\
\midrule
CFM--Geo
& 0.301 & 0.336 & 0.657
& 1.013 & 0.288 & 0.729\\
CFM--Lin
& 0.539 & 0.366 & 0.728
& 3.393 & 0.268 & 0.729\\
MF--Geo
& 0.370 & 0.343 & 0.643
& 1.846 & 0.261 & 0.746\\
MF--Lin
& 0.549 & 0.349 & 0.689
& 3.946 & 0.363 & 0.701\\
SM
& \textbf{0.190} & 0.296 & 0.735
& 1.191 & 0.213 & 0.763\\
LIFT--Geo
& 0.247 & 0.223 & 0.705
& \textbf{0.936} & 0.211 & 0.727\\
LIFT--Lin
& 0.269 & \textbf{0.119} & \textbf{0.803}
& 0.959 & \textbf{0.191} & \textbf{0.795}\\
\bottomrule
\end{tabular}
\end{table}

\begin{figure}[htbp]
\centering
    \hspace{-50 pt}
    \begin{subfigure}[b]{0.45\linewidth}
        \centering
        \includegraphics[width=\linewidth]{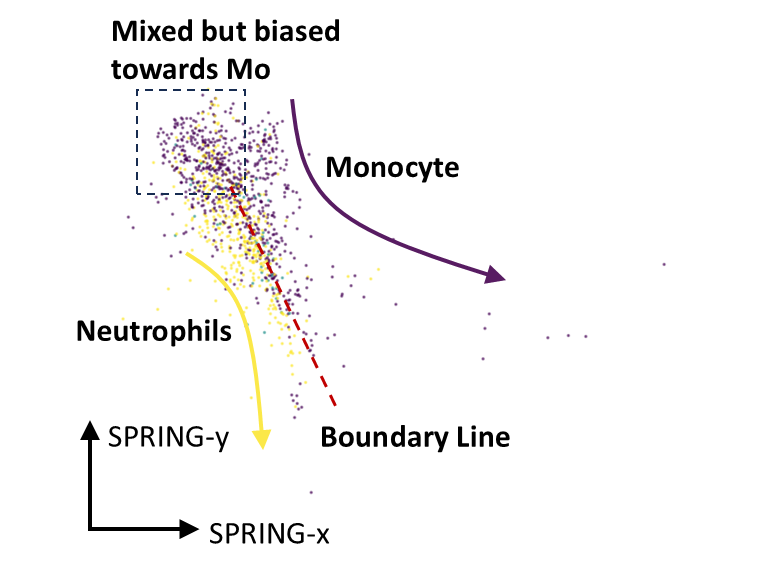}
        \caption{Lineage reference fate information}
        \label{mouse:sub1}
    \end{subfigure}
    \hspace{-75 pt}
    \begin{subfigure}[b]{0.45\linewidth}
        \centering
        \includegraphics[width=\linewidth]{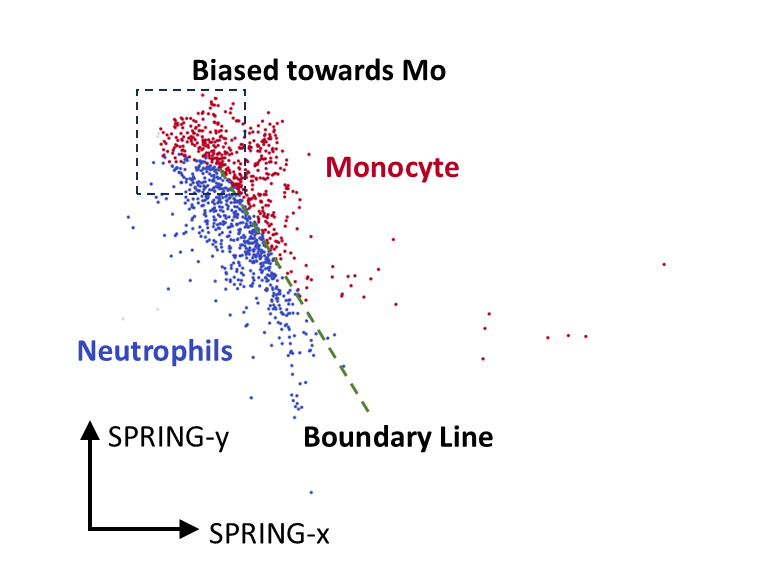}
        \caption{Hard Division}
        \label{mouse:sub2}
    \end{subfigure}
    \hspace{-75 pt}
    \begin{subfigure}[b]{0.45\linewidth}
        \centering
        \includegraphics[width=\linewidth]{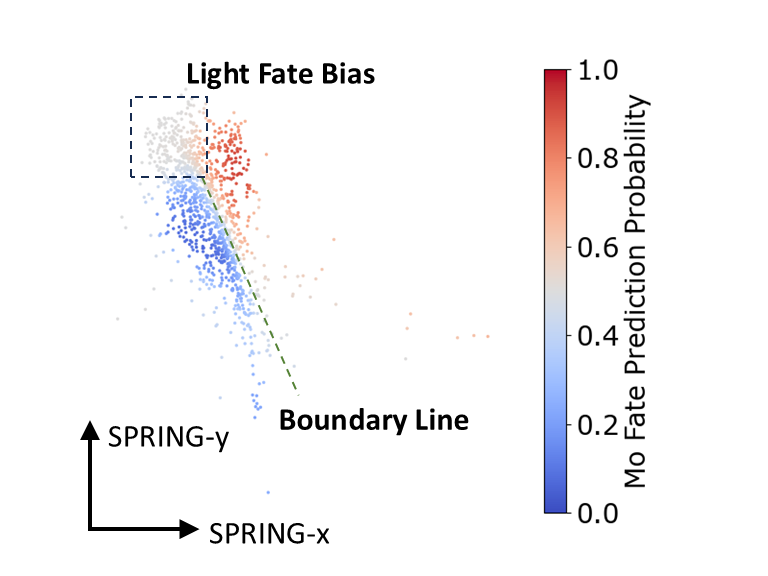}
        \caption{Soft Division}
        \label{mouse:sub3}
    \end{subfigure}
    \hspace{-60 pt}
\caption{Representative fate prediction in mouse hematopoiesis.
(a) Lineage-derived reference fate information.
(b) Hard fate assignments.
(c) Mo fate probabilities predicted by LIFT.
The highlighted region contains mixed Mo and Neu clonal outcomes
with a bias toward Mo. LIFT represents this region through
intermediate fate probabilities rather than exclusively binary
assignments.}
\label{fig:hematopoietic_fates}
\end{figure}

\paragraph{LIFT improves fate prediction with lineage information}
Table~\ref{tab:real_existing_results} shows that LIFT-Lin can achieve
the lowest MeanPE and highest correlation on both datasets.
Compared with LIFT-Geo, incorporating lineage information improves
fate-probability prediction even though the population Wasserstein
distance does not decrease. Supplying lineage-informed couplings
to deterministic CFM or MF does not produce comparable probability
accuracy. These results indicate that
accurate fate prediction requires pairing
lineage-informed couplings with a
stochastic generator capable of
representing one-to-many transitions.

\paragraph{LIFT captures graded fate bias in hematopoiesis}
Figure~\ref{fig:hematopoietic_fates} presents mouse hematopoiesis
as a representative example. In the highlighted region, the
lineage-derived reference contains mixed Mo and Neu outcomes with
a preference toward Mo. Hard assignments replace this mixture with
a sharp partition, whereas LIFT retains intermediate probabilities
with a modest Mo bias. The resulting probability map therefore
captures graded lineage-associated fate bias. This qualitative
agreement complements the quantitative results across both datasets.

\section{Discussion and Conclusion}
\label{sec:discussion_conclusion}

Lineage tracing links molecular states to clonal ancestry, but does not generally reveal unique cell-to-cell pairings across destructively sampled populations. This motivates a conditional-generation problem: the same observed state may be compatible with multiple future outcomes, whereas a deterministic state map assigns it a single endpoint. Consequently, reproducing the terminal population distribution alone does not ensure accurate fate probabilities. A model must also represent the one-to-many relationships supported by the available lineage information.

LIFT addresses this mismatch through a variational latent representation of conditional variability. An encoder learns latent distributions from endpoint pairs, a conditional prior predicts them from the initial state, and a Mean Flow decoder generates alternative outcomes without numerical integration. Under the stated regularity and prior-scale assumptions, we bound the true conditional distribution error by the discrepancy in the training coupling, decoder reconstruction error, and latent training-sampling mismatch. The analysis further connects these errors to fate-probability estimation, providing a theoretical basis for the conditional prediction objective.

The experiments show that LIFT combines accurate branching-probability prediction with competitive population distribution reconstruction and efficient one-step sampling. In mouse hematopoiesis and direct reprogramming, lineage-informed training improves fate prediction relative to the geometric-coupling variant. The hematopoietic example further illustrates how graded predictions represent mixed clonal outcomes instead of imposing hard assignments. Together, these results support the use of LIFT for efficient prediction of differentiation and reprogramming probabilities from molecular states and lineage information.

The current framework nevertheless has several limitations. Its generative dynamics do not explicitly model unbalanced effects such as proliferation and cell death, and its predictions remain dependent on the supplied coupling and chosen state representation. Endpoint constraints also do not uniquely identify intermediate biological paths or the sources of conditional variability. Future work will extend the framework to unbalanced dynamics, incorporate additional observation times and biological constraints. These extensions could improve both the biological fidelity of generated trajectories and the reliability of fate predictions.

\appendix

\section{Mathematical and Computational Background}
\label{app:background}

This appendix introduces the concepts connecting barcode-resolved observations, conditional generation, and fate-probability estimation.

\paragraph{Probabilistic modeling of cell dynamics}
A probability measure $\mu_t$ describes the relative abundance of
cell states at time $t$. Population evolution can therefore be modeled
as a continuous family of distributions connecting $\mu_0$ and $\mu_1$,
rather than as a single representative trajectory. Waddington-OT infers
probabilistic couplings between observed populations, accounting for
proliferation and death \cite{Schiebinger2019}. TrajectoryNet learns
continuous paths using neural ordinary differential equations and
optimal-transport regularization \cite{Tong2020}, while PRESCIENT
models population evolution through stochastic dynamics
\cite{Yeo2021}. These approaches connect snapshot observations through
assumed dynamics. A coupling additionally specifies how initial and
terminal states are paired; lineage information constrains these
relationships \cite{Forrow2021}. The conditional law
$K_\star(x,\cdot)$ then describes the possible outcomes of a fixed
initial state, which are not determined by the endpoint population
measures alone.

\paragraph{Variational inference and variational autoencoders}
Latent-variable models generate observations through auxiliary variables
whose values are not directly measured. Inferring their posterior
distribution is often computationally difficult. Variational inference
replaces this posterior with a tractable approximation and optimizes a
lower bound on the data log likelihood. Variational autoencoders (VAEs)
implement this idea using neural encoder and decoder networks
\cite{Kingma2014VAE,Kingma2019VAE}. Their objective balances
reconstruction against a Kullback--Leibler (KL) divergence penalty,
which compares the encoder distribution with the prior. In conditional
generation \cite{Sohn2015CVAE}, the model also receives an input $x$.
LIFT uses $q_\phi(z\mid x,y)$ during training and samples
$p_\eta(z\mid x)$ at inference. Different latent samples allow the
same initial state to generate different outcomes. The latent variable
represents conditional variability, not an automatically identified
biological mechanism.

\paragraph{Flow matching and Mean Flow}
Flow matching trains a velocity field by regression against velocities
of easily sampled reference paths \cite{Lipman2023FM,Tong2024CFM}.
For the linear path $X_t=(1-t)X+tY$, the reference velocity is $Y-X$.
Training therefore avoids solving the learned differential equation,
although generation with an FM decoder still requires numerical
integration. Mean Flow instead learns velocity averaged over a time
interval \cite{Geng2025MeanFlow}. For $r<t$, its average velocity
along a trajectory satisfies
\[
u(x_r,z;r,t)=\frac{x_t-x_r}{t-r},
\qquad
x_t=x_r+(t-r)u(x_r,z;r,t).
\]
Here, the average is over time along a trajectory, not over alternative
fates. Learning this quantity allows LIFT to predict each requested
state directly, with a latent code held fixed along a sampled path.

\paragraph{Wasserstein distance}
Wasserstein distance compares distributions by the displacement required
to move probability mass from one to the other
\cite{Peyre2019OT}. For probability measures $P,Q$ with finite second
moments,
\[
W_2^2(P,Q)
=
\inf_{\lambda\in\Pi(P,Q)}
\int\|a-b\|^2\,\lambda(\mathrm da,\mathrm db),
\]
where $\Pi(P,Q)$ contains all pairings with marginals $P$ and $Q$.
On finite cell populations, these pairings are nonnegative matrices
with prescribed row and column sums. Unlike comparing population
means, this distance measures differences in the locations and spread
of probability mass. Our conditional error applies it separately at
each initial state and averages the squared distances, distinguishing
correct future-state distributions from terminal population matching.

\paragraph{Further reading}

\emph{Biological observations and trajectory inference.}
Biddy et al.\ \cite{Biddy2018} introduce the CellTagging reprogramming
application; Weinreb et al.\ \cite{Weinreb2020} connect hematopoietic
state and fate using LARRY. Saelens et al.\ \cite{Saelens2019} compare
trajectory-inference methods and their assumptions. 

\emph{Lineage-informed endpoint relationships.}
LineageOT \cite{Forrow2021} explains how lineage trees inform transport
couplings, while CoSpar \cite{Wang2022} combines sparse clonal
observations with state-space structure to infer transition maps.

\emph{Latent generation and flow-based computation.}
Kingma and Welling \cite{Kingma2014VAE} and Sohn et al.\
\cite{Sohn2015CVAE} develop variational and conditional generative
models. Kingma and Welling also provide a tutorial on VAEs and variational training
\cite{Kingma2019VAE}. Lipman et al.\ \cite{Lipman2023FM} introduce flow-matching
training, and Geng et al.\ \cite{Geng2025MeanFlow} explain the
average-velocity formulation underlying Mean Flow.

\emph{Optimal transport:}
Peyr\'e and Cuturi provide a computational introduction to couplings,
Wasserstein distances, and their applications \cite{Peyre2019OT}.

\emph{Finite-sample probability estimates.}
Hoeffding \cite{Hoeffding1963} provides the concentration inequality
used to bound the sampling error of the estimated fate probabilities.

\section{Declarations}

\subsection*{Authorship and Contributorship}

All authors have made substantial intellectual contributions to the study conception, execution, and design of the work. All authors have read and approved the final manuscript. In addition, the following contributions occurred: Conceptualization: B. Yang; Methodology: B. Yang, J. Ying, and Y. Wang; Formal analysis and investigation: B. Yang, J. Ying, and Y. Wang; Writing -- original draft preparation: B. Yang; Writing -- review and editing: P. Zhou and L. Zhang; Funding acquisition: L. Zhang; Resources: L. Zhang; Supervision: P. Zhou and L. Zhang.

\subsection*{Conflicts of Interest}

The authors declare there are no conflicts of interest

\subsection*{Data \& Code Availability}

The datasets analyzed in this study are derived from previously published lineage-tracing studies, including the LARRY hematopoiesis dataset\cite{Weinreb2020} and the CellTagging direct reprogramming dataset\cite{Biddy2018}. The original datasets are available from the sources cited in the corresponding publications.

The code used to generate the results presented in this study are publicly available in \url{https://github.com/van-genshin-wander/LIFT}.

\subsection*{Ethics}

This study involves computational analyses of previously published and publicly available single-cell lineage-tracing datasets. No new experiments involving human participants or animals were conducted as part of this study.

\subsection*{Funding}

L. Zhang was supported by the National Natural Science Foundation of China (Nos. 12225102, T2321001, and 12288101) and the National Key Research and Development Program of China (No. 2024YFA0919500).

\subsection*{Use of AI Tools}

AI tools were used only to assist with language translation, editing, and polishing during the preparation of this manuscript. All authors have checked and verified the accuracy of all parts of this work and assume responsibility for all content.

\clearpage
\renewcommand{\siampretitle}{Supplementary Materials: }
\renewcommand{\siamprelabel}{SM}
\setcounter{page}{1}
\setcounter{section}{0}
\setcounter{subsection}{0}
\setcounter{subsubsection}{0}
\setcounter{equation}{0}
\setcounter{figure}{0}
\setcounter{table}{0}
\setcounter{theorem}{0}
\makeatletter
\renewcommand\section{\@startsection{section}{1}{.25in}%
                                  {1.3ex \@plus .5ex \@minus .2ex}%
                                  {-.5em \@plus -.1em}%
                                  {\reset@font\color{header1}\normalsize\HLtext}}
\makeatother

\renewcommand{\thesection}{SM\arabic{section}}
\renewcommand{\thesubsection}{\thesection.\arabic{subsection}}
\renewcommand{\thesubsubsection}{\thesubsection.\arabic{subsubsection}}
\headers{LIFT}{B. Yang, J. Ying, Y. Wang, P. Zhou and L. Zhang}

\title{LIFT: Latent Flow Matching for Probabilistic Cell Fate Dynamics with Single Cell Snapshots}
\author{Bowen Yang \and Junda Ying \and Yuxuan Wang \and Peijie Zhou \and Lei Zhang}
\LIFTsuppmaketitle
\thispagestyle{plain}

\section{Proof of Proposition\ref{prop:deterministic_conditional_error}
}
\label{app:prop_proof}
\begin{proof}
There is only one coupling between $K_\star(x,\cdot)$ and
$\delta_{T(x)}$. Hence, by definition, 
\[
W_2^2\!\left(
K_\star(x,\cdot),\delta_{T(x)}
\right)
=
\int\|y-T(x)\|^2K_\star(x,\mathrm dy).
\]
Integrating over the source marginal yields
\[
\mathcal E_\star^2(T)
=
\mathbb E_{\pi_\star}\|Y-T(X)\|^2.
\]

Decompose
\[
Y-T(X)
=
Y-m_\star(X)+m_\star(X)-T(X).
\]
The cross term vanishes because
\[
\mathbb E_{\pi_\star}[Y-m_\star(X)\mid X]=0.
\]
Therefore,
\[
\begin{aligned}
\mathbb E_{\pi_\star}\|Y-T(X)\|^2
={}&
\mathbb E_{\pi_\star}\|Y-m_\star(X)\|^2
+
\mathbb E_{\pi_\star}\|m_\star(X)-T(X)\|^2.
\end{aligned}
\]
The first term equals
$\int\operatorname{tr}\Sigma_\star(x)\,\mu_0(\mathrm dx)$,
proving \eqref{eq:deterministic_error_decomposition}.
The second term is nonnegative and vanishes for $T=m_\star$, which proves
\eqref{eq:irreducible_deterministic_error}.

Finally, a deterministic realization has conditional law
$\delta_{T(x)}$ almost everywhere. Conversely, if
$K_\star(x,\cdot)$ is a Dirac measure almost everywhere, its point
of concentration is $m_\star(x)$, and
\[
\pi_\star=(\operatorname{Id},m_\star)_\#\mu_0.
\]
\end{proof}

\section{Proofs of the Error Estimates}
\label{app:error_proofs}

We provide the proofs of the results in
Section~\ref{sec:error_analysis}. Throughout the appendix, the
model parameters are fixed, and
\[
Q_\phi(\mathrm dx,\mathrm dy,\mathrm dz)
=
\pi(\mathrm dx,\mathrm dy)q_\phi(\mathrm dz\mid x,y).
\]
All statements about conditional distributions are understood to
hold for $\mu_0$-almost every initial state. We write
\[
X_t=(1-t)X+tY,
\qquad
V=Y-X,
\]
for the reference path and its velocity. Norms without a subscript
are Euclidean norms. The finite second moments of the endpoint
measures imply that $V\in L^2(Q_\phi;\mathbb R^d)$.

\subsection{Proof of Theorem~\ref{thm:fm_reconstruction_control}}
\label{app:fm_proof}

\begin{proof}
Fix a triple $(X,Y,Z)$ for which the generated ODE has a solution
and the reference-path regression residual is square-integrable
in time. These properties hold $Q_\phi$-almost surely under the
theorem's assumptions. Let
\[
\dot{\widehat X}_t
=
v_\theta(\widehat X_t,Z,t),
\qquad
\widehat X_0=X,
\]
and define
\[
e_t=\widehat X_t-X_t,
\qquad
a_t=v_\theta(X_t,Z,t)-V.
\]
The difference between the generated and reference paths satisfies
\[
\dot e_t
=
v_\theta(\widehat X_t,Z,t)
-
v_\theta(X_t,Z,t)
+a_t,
\qquad e_0=0.
\]
The state-space Lipschitz assumption gives the integral inequality
\[
\|e_t\|
\le
\int_0^t\left(L_x\|e_s\|+\|a_s\|\right)\,\mathrm ds.
\]
Gronwall's inequality therefore yields
\begin{equation}
\|e_t\|
\le
\int_0^t e^{L_x(t-s)}\|a_s\|\,\mathrm ds.
\label{eq:app_fm_pathwise_stability}
\end{equation}
At $t=1$, the Cauchy--Schwarz inequality implies
\[
\begin{aligned}
\|\widehat X_1-Y\|^2
&\le
\left(\int_0^1 e^{L_x(1-s)}\|a_s\|\,\mathrm ds\right)^2\\
&\le
\left(\int_0^1e^{2L_x(1-s)}\,\mathrm ds\right)
\left(\int_0^1\|a_s\|^2\,\mathrm ds\right).
\end{aligned}
\]
The first factor equals
\[
\int_0^1e^{2L_x(1-s)}\,\mathrm ds
=
\begin{cases}
\dfrac{e^{2L_x}-1}{2L_x},&L_x>0,\\[2mm]
1,&L_x=0.
\end{cases}
\]
Taking expectation under $Q_\phi$ gives
\[
D_\pi^{\mathrm{FM}}
\le
C_{L_x}\,
\mathbb E_{Q_\phi}
\int_0^1\|v_\theta(X_s,Z,s)-V\|^2\,\mathrm ds
=
C_{L_x}\mathcal L_{\mathrm{FM}}.
\]
\end{proof}

\subsection{Proof of Theorem~\ref{thm:mf_reconstruction_control}}
\label{app:mf_proof}

The proof first relates the Mean Flow residual to a direct prediction error at each terminal time. The time-regularity assumption then converts an integrated error bound into control at $t=1$.

\begin{lemma}[Integrated Mean Flow residual]
\label{lem:app_mf_residual_identity}
Under the absolute-continuity assumption of
Theorem~\ref{thm:mf_reconstruction_control}, define
\[
G_{\theta,t}(X,Z)=X+t\,u_\theta(X,Z;0,t).
\]
Then, for every $t\in(0,1]$,
\begin{equation}
G_{\theta,t}(X,Z)-X_t
=
\int_0^t\mathscr R_{r,t}\,\mathrm dr
\qquad Q_\phi\text{-almost surely}.
\label{eq:app_mf_residual_identity}
\end{equation}
In particular, if
\[
A(t)
=
\mathbb E_{Q_\phi}
\left[\frac1t\int_0^t\|\mathscr R_{r,t}\|^2\,\mathrm dr\right],
\]
then
\begin{equation}
\mathbb E_{Q_\phi}
\|u_\theta(X,Z;0,t)-V\|^2
\le A(t),
\label{eq:app_mf_fixed_time_bound}
\end{equation}
and
\begin{equation}
\mathbb E_{Q_\phi}
\|G_{\theta,t}(X,Z)-X_t\|^2
\le t^2A(t).
\label{eq:app_mf_intermediate_bound}
\end{equation}
\end{lemma}

\begin{proof}
Fix $t\in(0,1]$ and a triple for which the assumed absolute
continuity holds. Write
\[
U_t(r)=u_\theta(X_r,Z;r,t),
\qquad
H_t(r)=X_r+(t-r)U_t(r).
\]
The chain rule along the reference path gives
\[
U_t'(r)
=
\partial_r u_\theta(X_r,Z;r,t)
+
D_xu_\theta(X_r,Z;r,t)V
\]
for almost every $r$. Consequently,
\[
H_t'(r)
=
V-U_t(r)+(t-r)U_t'(r)
=
-\mathscr R_{r,t}.
\]
Since $H_t$ is absolutely continuous and
\[
H_t(0)=G_{\theta,t}(X,Z),
\qquad
H_t(t)=X_t,
\]
the fundamental theorem of calculus proves
\eqref{eq:app_mf_residual_identity}.

Using $X_t=X+tV$ and dividing by $t$ yields
\[
u_\theta(X,Z;0,t)-V
=
\frac1t\int_0^t\mathscr R_{r,t}\,\mathrm dr.
\]
The Cauchy--Schwarz inequality gives
\[
\|u_\theta(X,Z;0,t)-V\|^2
\le
\frac1t\int_0^t\|\mathscr R_{r,t}\|^2\,\mathrm dr.
\]
Taking expectation proves \eqref{eq:app_mf_fixed_time_bound}.
Multiplication by $t^2$ gives
\eqref{eq:app_mf_intermediate_bound}.
These inequalities remain valid with an infinite right-hand side
at exceptional times.
\end{proof}

\begin{proof}[Proof of Theorem~\ref{thm:mf_reconstruction_control}]
Let
\[
\mathcal H=L^2(Q_\phi;\mathbb R^d),
\qquad
g(t)=u_\theta(X,Z;0,t)-V\in\mathcal H,
\qquad
J=\mathcal L_{\mathrm{MF}}.
\]
Because $t$ is uniform on $[0,1]$ and $r\mid t$ is uniform on
$[0,t]$, Tonelli's theorem and
Lemma~\ref{lem:app_mf_residual_identity} imply
\begin{equation}
\int_0^1\|g(t)\|_{\mathcal H}^2\,\mathrm dt
\le
\int_0^1 A(t)\,\mathrm dt
=J.
\label{eq:app_mf_integrated_error}
\end{equation}
The value assigned to the conditional time distribution at $t=0$
has no effect on this integral.

The time-regularity assumption gives
\[
\|g(t)-g(s)\|_{\mathcal H}
\le L_T|t-s|.
\]
Set
\[
a=\|g(1)\|_{\mathcal H}
=\sqrt{D_\pi^{\mathrm{MF}}}.
\]
The reverse triangle inequality yields
\begin{equation}
\|g(t)\|_{\mathcal H}
\ge
\max\{0,a-L_T(1-t)\}.
\label{eq:app_mf_terminal_lower_envelope}
\end{equation}
If $a=0$, the conclusion is immediate. If $L_T=0$, then $g$ is
constant, and \eqref{eq:app_mf_integrated_error} gives
$a^2\le J$, which is stronger than the claimed estimate.
It remains to consider $a>0$ and $L_T>0$.

If $a<L_T$, integration over the interval where the lower bound
in \eqref{eq:app_mf_terminal_lower_envelope} is positive gives
\[
\begin{aligned}
J
&\ge
\int_{1-a/L_T}^1
\bigl(a-L_T(1-t)\bigr)^2\,\mathrm dt\\
&=\frac{a^3}{3L_T}.
\end{aligned}
\]
Hence $a^2\le(3L_TJ)^{2/3}$.

If $a\ge L_T$, the lower bound is nonnegative throughout
$[0,1]$, and
\[
\begin{aligned}
J
&\ge
\int_0^1\bigl(a-L_T(1-t)\bigr)^2\,\mathrm dt\\
&=a^2-aL_T+\frac{L_T^2}{3}
\ge\frac{a^2}{3}.
\end{aligned}
\]
The last inequality follows from
\[
a^2-aL_T+\frac{L_T^2}{3}-\frac{a^2}{3}
=
\frac{(a-L_T)(2a-L_T)}{3}\ge0.
\]
Thus $a^2\le3J$ in this case. Combining the cases proves
\[
D_\pi^{\mathrm{MF}}
\le
\max\{3\mathcal L_{\mathrm{MF}},
(3L_T\mathcal L_{\mathrm{MF}})^{2/3}\}.
\]
\end{proof}

\paragraph{Sharpness of the small-loss exponent.}
The exponent $2/3$ is attained by a simple decoder satisfying the
stated assumptions. Let $X=Y=0$ and, for fixed $L>0$ and
$0<a<L$, take a scalar decoder independent of its state, latent,
and starting-time arguments:
\[
u_\theta(\xi,z;r,t)
=
\bigl(a-L(1-t)\bigr)_+,
\qquad b_+=\max\{b,0\}.
\]
Its reference-path derivative with respect to $r$ vanishes, so
$\mathscr R_{r,t}=u_\theta(t)$. Therefore,
\[
\mathcal L_{\mathrm{MF}}=\frac{a^3}{3L},
\qquad
D_\pi^{\mathrm{MF}}=a^2
=(3L\mathcal L_{\mathrm{MF}})^{2/3}.
\]
As $a\downarrow0$ with $L$ fixed, no uniformly stronger small-loss
power can replace $2/3$ under these assumptions alone.

\subsection{Proof of Theorem~\ref{thm:latent_true_conditional_error}}
\label{app:total_error_proof}

We first record an elementary Gaussian comparison. The proof uses an explicit coupling and will therefore also apply after mixing encoder distributions over terminal states.

\begin{lemma}[Gaussian coupling and relative entropy]
\label{lem:app_gaussian_comparison}
Let
\[
q=\mathcal N(m_q,\operatorname{diag}(s_q^2)),
\qquad
p=\mathcal N(m_p,\operatorname{diag}(s_p^2))
\]
be diagonal Gaussians on $\mathbb R^{d_z}$ with strictly positive
standard deviations. If $s_{p,j}\le S$ for every coordinate, then
\begin{equation}
\begin{aligned}
W_2^2(q,p)
&\le
\|m_q-m_p\|^2+\|s_q-s_p\|^2\\
&\le
2S^2D_{\mathrm{KL}}(q\|p).
\end{aligned}
\label{eq:app_gaussian_comparison}
\end{equation}
\end{lemma}

\begin{proof}
For a common standard Gaussian vector
$\epsilon\sim\mathcal N(0,I_{d_z})$, set
\[
Z=m_q+s_q\odot\epsilon,
\qquad
Z'=m_p+s_p\odot\epsilon.
\]
These variables have marginals $q$ and $p$, and their coupling cost
is
\[
\mathbb E\|Z-Z'\|^2
=
\|m_q-m_p\|^2+\|s_q-s_p\|^2.
\]
This proves the first inequality.

For the second, set $\alpha_j=s_{q,j}/s_{p,j}>0$.
The Gaussian relative entropy is
\[
2D_{\mathrm{KL}}(q\|p)
=
\sum_{j=1}^{d_z}
\left[
\frac{(m_{q,j}-m_{p,j})^2}{s_{p,j}^2}
+
\alpha_j^2-1-2\log\alpha_j
\right].
\]
For every $\alpha>0$,
\[
\alpha^2-1-2\log\alpha-(\alpha-1)^2
=
2(\alpha-1-\log\alpha)\ge0.
\]
Consequently, for each coordinate,
\[
\begin{aligned}
&(m_{q,j}-m_{p,j})^2+(s_{q,j}-s_{p,j})^2\\
&\quad\le
s_{p,j}^2
\left[
\frac{(m_{q,j}-m_{p,j})^2}{s_{p,j}^2}
+
\alpha_j^2-1-2\log\alpha_j
\right]\\
&\quad\le
S^2
\left[
\frac{(m_{q,j}-m_{p,j})^2}{s_{p,j}^2}
+
\alpha_j^2-1-2\log\alpha_j
\right].
\end{aligned}
\]
Summing over coordinates proves the result.
\end{proof}

\begin{proof}[Proof of Theorem~\ref{thm:latent_true_conditional_error}]
We separate reconstruction under encoder samples from the change
of latent distribution at inference.

\paragraph{Step 1: Reconstruction under the aggregated posterior.}
Define
\[
\widetilde K_1(x,\cdot)
=
F_\theta(x,\cdot)_\#\overline q_\phi(\cdot\mid x)
\]
and the conditional reconstruction cost
\[
d_\pi(x)
=
\int\!\int
\|y-F_\theta(x,z)\|^2
q_\phi(\mathrm dz\mid x,y)K_\pi(x,\mathrm dy).
\]
Then $\int d_\pi(x)\,\mu_0(\mathrm dx)=D_\pi$.
For fixed $x$, sample $Y\sim K_\pi(x,\cdot)$ and then
$Z\sim q_\phi(\cdot\mid x,Y)$.
The pair $(Y,F_\theta(x,Z))$ is a coupling of
$K_\pi(x,\cdot)$ and $\widetilde K_1(x,\cdot)$, because the
marginal law of $Z$ is $\overline q_\phi(\cdot\mid x)$.
It follows that
\begin{equation}
W_2\!\left(K_\pi(x,\cdot),\widetilde K_1(x,\cdot)\right)
\le\sqrt{d_\pi(x)}.
\label{eq:app_reconstruction_kernel_bound}
\end{equation}

\paragraph{Step 2: Replacing the aggregated posterior by the prior.}
Fix $x$ and take any coupling of
$\overline q_\phi(\cdot\mid x)$ and $p_\eta(\cdot\mid x)$.
Pushing both components through $F_\theta(x,\cdot)$ gives a
coupling of $\widetilde K_1(x,\cdot)$ and
$\widehat K_1(x,\cdot)$.
By the latent-variable Lipschitz assumption, its squared cost is
at most $L_Z^2$ times the original cost.
Taking the infimum over latent couplings yields
\begin{equation}
W_2\!\left(\widetilde K_1(x,\cdot),\widehat K_1(x,\cdot)\right)
\le
L_ZW_2\!\left(
\overline q_\phi(\cdot\mid x),p_\eta(\cdot\mid x)
\right).
\label{eq:app_prior_pushforward_bound}
\end{equation}

The Wasserstein triangle inequality, followed by
\eqref{eq:app_reconstruction_kernel_bound} and
\eqref{eq:app_prior_pushforward_bound}, gives
\[
\begin{aligned}
&W_2\!\left(K_\star(x,\cdot),\widehat K_1(x,\cdot)\right)\\
&\quad\le
W_2\!\left(K_\star(x,\cdot),K_\pi(x,\cdot)\right)
+\sqrt{d_\pi(x)}\\
&\qquad\quad+
L_ZW_2\!\left(
\overline q_\phi(\cdot\mid x),p_\eta(\cdot\mid x)
\right).
\end{aligned}
\]
Taking the $L^2(\mu_0)$ norm and applying Minkowski's inequality
proves
\begin{equation}
\mathcal E_\star(\widehat K_1)
\le
\varepsilon_{\mathrm{cpl}}+\sqrt{D_\pi}+L_Z\Delta_Z.
\label{eq:app_total_error_decomposition}
\end{equation}
In particular, this is a triangle-inequality estimate, not an
orthogonal decomposition of squared errors.

\paragraph{Step 3: Gaussian control of the latent discrepancy.}
Fix $x$ and sample $Y\sim K_\pi(x,\cdot)$ and an independent
$\epsilon\sim\mathcal N(0,I_{d_z})$. Define
\[
\begin{aligned}
Z&=m_\phi(x,Y)+s_\phi(x,Y)\odot\epsilon,\\
Z'&=m_\eta(x)+s_\eta(x)\odot\epsilon.
\end{aligned}
\]
The marginal law of $Z$ is the aggregated posterior, and the
marginal law of $Z'$ is the prior. Hence,
\[
\begin{aligned}
&W_2^2\!\left(
\overline q_\phi(\cdot\mid x),p_\eta(\cdot\mid x)
\right)\\
&\quad\le
\int
\left[
\|m_\phi(x,y)-m_\eta(x)\|^2
+
\|s_\phi(x,y)-s_\eta(x)\|^2
\right]K_\pi(x,\mathrm dy)\\
&\quad\le
2S^2\int
D_{\mathrm{KL}}\!\left(
q_\phi(\cdot\mid x,y)\|p_\eta(\cdot\mid x)
\right)K_\pi(x,\mathrm dy),
\end{aligned}
\]
where the second inequality follows from
Lemma~\ref{lem:app_gaussian_comparison}.
Integration with respect to $\mu_0$ gives
\[
\Delta_Z^2\le2S^2\mathcal R_{\mathrm{KL}}.
\]
This argument does not assume that the aggregated posterior is
Gaussian; it explicitly couples the mixture to the Gaussian prior.

Finally, substitute this estimate and
\[
D_\pi\le\Psi_{\mathrm{dec}}(\mathcal L_{\mathrm{dec}})
\]
from the appropriate reconstruction theorem into
\eqref{eq:app_total_error_decomposition}. This yields
\[
\mathcal E_\star(\widehat K_1)
\le
\varepsilon_{\mathrm{cpl}}
+
\sqrt{\Psi_{\mathrm{dec}}(\mathcal L_{\mathrm{dec}})}
+
\sqrt{2}\,L_ZS\sqrt{\mathcal R_{\mathrm{KL}}},
\]
as claimed.
\end{proof}

\subsection{Proof of Proposition~\ref{prop:fate_probability_stability}}
\label{app:fate_probability_proof}

\begin{proof}
Fix a fate $k$ and a radius $\rho>0$. Recall
\[
\Gamma=\bigcup_{j=1}^K\partial B_j,
\qquad
\Gamma_\rho=\{y:\operatorname{dist}(y,\Gamma)\le\rho\}.
\]
For any $y,\widehat y\in\mathbb R^d$,
\begin{equation}
\left|\mathbf 1_{B_k}(y)-\mathbf 1_{B_k}(\widehat y)\right|
\le
\mathbf 1_{\Gamma_\rho}(y)
+
\mathbf 1_{\{\|y-\widehat y\|>\rho\}}.
\label{eq:app_label_change_bound}
\end{equation}
Indeed, if $y\notin\Gamma_\rho$ and
$\|y-\widehat y\|\le\rho$, the line segment joining the two
points does not intersect $\partial B_k$. Membership in $B_k$
cannot change along this segment, so the two indicators agree.
No smoothness of the boundary is required for this observation.

For fixed $x$, let $\lambda$ be any coupling of
$K_\star(x,\cdot)$ and $\widehat K_1(x,\cdot)$.
Using \eqref{eq:app_label_change_bound},
\[
\begin{aligned}
|p_k^\star(x)-\widehat p_k(x)|
&\le
\int
\left|\mathbf 1_{B_k}(y)-\mathbf 1_{B_k}(\widehat y)\right|
\lambda(\mathrm dy,\mathrm d\widehat y)\\
&\le
K_\star(x,\Gamma_\rho)
+
\lambda\{\|y-\widehat y\|>\rho\}\\
&\le
K_\star(x,\Gamma_\rho)
+
\frac1{\rho^2}
\int\|y-\widehat y\|^2
\lambda(\mathrm dy,\mathrm d\widehat y).
\end{aligned}
\]
Taking the infimum over $\lambda$ gives the pointwise estimate
\begin{equation}
|p_k^\star(x)-\widehat p_k(x)|
\le
K_\star(x,\Gamma_\rho)
+
\frac{W_2^2(K_\star(x,\cdot),\widehat K_1(x,\cdot))}{\rho^2}.
\label{eq:app_pointwise_fate_error}
\end{equation}
This pointwise argument does not require selecting optimal
couplings measurably as a function of $x$.

Integrating against $\mu_0$ and using the true terminal marginal,
\[
\int K_\star(x,\Gamma_\rho)\,\mu_0(\mathrm dx)
=
\mu_1(\Gamma_\rho),
\]
we obtain
\[
\|p_k^\star-\widehat p_k\|_{L^1(\mu_0)}
\le
\mu_1(\Gamma_\rho)
+
\frac{\mathcal E_\star^2(\widehat K_1)}{\rho^2}.
\]
The right-hand side is independent of $k$, so the same estimate
holds after taking the maximum over fates.
Theorem~\ref{thm:latent_true_conditional_error} gives the further
bound obtained by replacing $\mathcal E_\star$ with
$\mathcal B_{\mathrm{dec}}$.

Finally, $\Gamma$ is closed because it is a finite union of
boundaries. Therefore $\Gamma_\rho\downarrow\Gamma$ as
$\rho\downarrow0$, and
\[
\mu_1(\Gamma_\rho)\longrightarrow\mu_1(\Gamma).
\]
If $\mu_1(\Gamma)=0$, first choose $\rho$ so that the boundary
mass is small, and then let the conditional reconstruction error
tend to zero with this $\rho$ fixed. This proves the stated
convergence of the average fate probabilities.
\end{proof}

\subsection{Proof of Monte-Carlo error bound}
\label{app:monte_carlo_proof}

\begin{proposition}[Monte Carlo estimation error]
\label{prop:mc_fate_error}
For each $a$ and $k$, the estimator
\begin{equation}
\widehat p_k^{(N)}(x_a)
=
\frac1N\sum_{\ell=1}^N
\mathbf 1_{B_k}(\widehat Y_{a,\ell}).
\label{eq:mc_fate_probability}
\end{equation} is unbiased for
$\widehat p_k(x_a)$ and satisfies
\begin{equation}
\mathbb E_{\mathrm{MC}}
\left|
\widehat p_k^{(N)}(x_a)-\widehat p_k(x_a)
\right|^2
=
\frac{
\widehat p_k(x_a)(1-\widehat p_k(x_a))
}{N}
\le
\frac1{4N}.
\label{eq:mc_fate_mse}
\end{equation}
Moreover, for any $\delta\in(0,1)$, with probability at least
$1-\delta$,
\begin{equation}
\max_{\substack{1\le a\le n\\1\le k\le K}}
\left|
\widehat p_k^{(N)}(x_a)-\widehat p_k(x_a)
\right|
\le
\epsilon_{\mathrm{MC}},
\label{eq:mc_uniform_fate_bound}
\end{equation}
where
\begin{equation}
\epsilon_{\mathrm{MC}}
=
\sqrt{\frac{\log(2nK/\delta)}{2N}}.
\label{eq:mc_error_radius}
\end{equation}
\end{proposition}

\begin{proof}
Condition on the fitted model and the query states. For fixed
$a$ and $k$, set
\[
I_\ell=\mathbf 1_{B_k}(\widehat Y_{a,\ell}),
\qquad
p=\widehat p_k(x_a).
\]
The variables $I_1,\ldots,I_N$ are independent Bernoulli variables
with success probability $p$. Thus,
\[
\mathbb E_{\mathrm{MC}}\widehat p_k^{(N)}(x_a)=p,
\]
and independence gives
\[
\begin{aligned}
\mathbb E_{\mathrm{MC}}
\left|\widehat p_k^{(N)}(x_a)-p\right|^2
&=
\operatorname{Var}_{\mathrm{MC}}
\left(\frac1N\sum_{\ell=1}^NI_\ell\right)\\
&=
\frac{p(1-p)}{N}
\le\frac1{4N}.
\end{aligned}
\]

For completeness, we give the exponential-moment argument for
the required concentration bound. For a Bernoulli variable $I$
with success probability $p$, define
\[
H(\lambda)
=
\log\mathbb E e^{\lambda(I-p)}
=
\log(1-p+pe^\lambda)-p\lambda.
\]
Then $H(0)=H'(0)=0$, and
\[
H''(\lambda)
=
\vartheta_\lambda(1-\vartheta_\lambda)
\le\frac14,
\qquad
\vartheta_\lambda
=
\frac{pe^\lambda}{1-p+pe^\lambda}.
\]
Taylor's formula therefore implies
\[
\mathbb E e^{\lambda(I-p)}
\le e^{\lambda^2/8}
\qquad\text{for every }\lambda\in\mathbb R.
\]
By independence and exponential Markov's inequality, for
$\varepsilon>0$ and $\lambda>0$,
\[
\begin{aligned}
\mathbb P_{\mathrm{MC}}
\left(
\frac1N\sum_{\ell=1}^NI_\ell-p\ge\varepsilon
\right)
&\le
\exp\left(-N\lambda\varepsilon+\frac{N\lambda^2}{8}\right).
\end{aligned}
\]
Choosing $\lambda=4\varepsilon$ gives the upper-tail bound
$e^{-2N\varepsilon^2}$. Applying the same argument to the
negative centered sum gives
\begin{equation}
\mathbb P_{\mathrm{MC}}
\left(
\left|\widehat p_k^{(N)}(x_a)-\widehat p_k(x_a)\right|
>\varepsilon
\right)
\le 2e^{-2N\varepsilon^2}.
\label{eq:app_bernoulli_tail_bound}
\end{equation}
This is the Bernoulli form of Hoeffding's inequality.

A union bound over the $nK$ pairs $(a,k)$ gives
\[
\mathbb P_{\mathrm{MC}}
\left(
\max_{a,k}
\left|\widehat p_k^{(N)}(x_a)-\widehat p_k(x_a)\right|
>\varepsilon
\right)
\le
2nK e^{-2N\varepsilon^2}.
\]
Set
\[
\varepsilon
=
\epsilon_{\mathrm{MC}}
=
\sqrt{\frac{\log(2nK/\delta)}{2N}}.
\]
The right-hand side then equals $\delta$, proving the claimed
simultaneous bound. Independence between indicators of different
fates is not needed for this union-bound step.
\end{proof}

\subsection{Proof of Corollary~\ref{cor:end_to_end_fate_error}}
\label{app:end_to_end_fate_proof}

\begin{proof}
Let $\mathcal A_\delta$ be the event
\[
\max_{a,k}
\left|\widehat p_k^{(N)}(x_a)-\widehat p_k(x_a)\right|
\le\epsilon_{\mathrm{MC}}.
\]
By Proposition~\ref{prop:mc_fate_error},
$\mathbb P_{\mathrm{MC}}(\mathcal A_\delta)\ge1-\delta$.
On this event, for every fate $k$,
\[
\begin{aligned}
&\sum_{a=1}^n\omega_a
\left|\widehat p_k^{(N)}(x_a)-p_k^\star(x_a)\right|\\
&\quad\le
\sum_{a=1}^n\omega_a
\left|\widehat p_k(x_a)-p_k^\star(x_a)\right|
+
\sum_{a=1}^n\omega_a
\left|\widehat p_k^{(N)}(x_a)-\widehat p_k(x_a)\right|\\
&\quad\le
\|\widehat p_k-p_k^\star\|_{L^1(\mu_0)}
+
\epsilon_{\mathrm{MC}},
\end{aligned}
\]
where we used
$\mu_0=\sum_a\omega_a\delta_{x_a}$ and $\sum_a\omega_a=1$.
Proposition~\ref{prop:fate_probability_stability} now implies
that, on the same event and for every $\rho>0$,
\[
\max_k\sum_{a=1}^n\omega_a
\left|\widehat p_k^{(N)}(x_a)-p_k^\star(x_a)\right|
\le
\mu_1(\Gamma_\rho)
+
\frac{\mathcal B_{\mathrm{dec}}^2}{\rho^2}
+
\epsilon_{\mathrm{MC}}.
\]
Taking the infimum over $\rho>0$ and substituting the expression
for $\epsilon_{\mathrm{MC}}$ proves
\[
\begin{aligned}
&\max_k\sum_{a=1}^n\omega_a
\left|\widehat p_k^{(N)}(x_a)-p_k^\star(x_a)\right|\\
&\quad\le
\inf_{\rho>0}
\left\{
\mu_1(\Gamma_\rho)
+
\frac{\mathcal B_{\mathrm{dec}}^2}{\rho^2}
\right\}
+
\sqrt{\frac{\log(2nK/\delta)}{2N}}.
\end{aligned}
\]
The event $\mathcal A_\delta$ does not depend on $\rho$.
Thus, taking this infimum introduces no additional probability
penalty.
\end{proof}

\providecommand{\expblank}{\mbox{\underline{\hspace{1.2em}}}}
\providecommand{\expnote}[1]{\par\smallskip\noindent
  {\small\textbf{[Draft note.] }#1}\par\smallskip}

\section{Additional Experimental Details and Results}
\label{sec:sm_experiments}

This section provides the implementation settings, numerical-integration comparisons, hyperparameter sensitivity results, and additional hematopoietic reconstruction scores. We also present qualitative comparisons of the learned fields, latent representations, and baseline fate maps. Population reconstruction and fate-probability accuracy are reported separately throughout.

\subsection{Experimental setup and evaluation protocol}
\label{subsec:experimental_setup}

\paragraph{Data and implementation}
The controlled benchmark is a two-dimensional branching system with
an analytically specified branch probability. The biological applications use
barcode-resolved hematopoietic data from LARRY
\cite{Weinreb2020} and direct-reprogramming data from CellTagging
\cite{Biddy2018}. Dataset sizes and the correspondence
between experimental and rescaled times are recorded in
Table~\ref{tab:dataset_overview}.

The reported LIFT implementation uses multilayer perceptrons with ReLU activations for the encoder, conditional prior, and decoder, with a two-dimensional latent variable in the principal experiments. Training uses Adam and the objective introduced in Section~\ref{subsec:training_sampling}. Our timing results were obtained on an Intel i7-13700 CPU. Dataset-specific network widths, optimization parameters, and sensitivity results are contained below.

\begin{table}[t]
\centering
\caption{Experimental datasets. Here we use low dimension representation of real datasets deduced from standard SPRING\cite{Weinreb2018SPRING} method.}
\label{tab:dataset_overview}
\small
\setlength{\tabcolsep}{4pt}
\begin{tabular}{lcccc}
\toprule
Dataset & Time interval & Source cells & Target cells\\
\midrule
Synthetic branching & $0\to1$ & 1,000 & 5,000\\
Hematopoiesis & Day 2$\to$Day 6 & 1144 & 6477\\
Reprogramming & Day 15$\to$Day 28 & 1897 & 5945\\
\bottomrule
\end{tabular}
\end{table}

\paragraph{Training couplings and comparison methods}
We compare geometric endpoint couplings with lineage-informed
couplings, denoted by $\pi_{\mathrm{geo}}$ and $\pi_{\mathrm{lin}}$.
The latter use shared barcodes to restrict the admissible source--target
relationships; the reprogramming experiment additionally uses OT
within each retained clone. During LIFT sampling, only the molecular
state is provided to the conditional prior. Barcode labels and terminal
states are not inputs to the sampling network.

The primary comparisons include deterministic CFM and Mean Flow, the score-matching baseline denoted SM. We also compare latent decoders sharing the encoder--prior construction: a static MLP, an FM decoder, and the Mean Flow decoder used by LIFT. These experiments distinguish changes in the coupling from changes in the generator.

\paragraph{Evaluation and data separation}
Population reconstruction is assessed by a Wasserstein distance in
the evaluation state representation. Conditional evaluation uses
analytic branch probabilities in the controlled benchmark and
lineage derived outcome information in the biological applications.
LIFT
probabilities are estimated using 1,000 independent prior samples
per initial state. The terminal fate assignment rule is shared across
methods and is specified by using the convex hull as fate region. Throughout evaluated datasets, convex hulls of different labeled fates do not intersect so this assignment is well-defined. As for evaluation, we randomly separate cells from 20\% barcodes as test set.

\subsection{LIFT implementation}
\label{subsec:sm_implementation}

The encoder, conditional prior, and Mean Flow decoder are multilayer perceptrons with ReLU activations. The encoder receives the concatenated endpoint pair $(x,y)$ and parameterizes a diagonal Gaussian latent posterior. The conditional prior parameterizes a diagonal Gaussian using $x$ alone. The decoder receives the state, latent variable, starting time, and terminal time, and returns the interval-average velocity. The networks are trained jointly with Adam. Table~\ref{tab:sm_implementation} lists the dataset-specific architecture and optimization settings; the real-data configuration is shared by the hematopoiesis and reprogramming experiments.

\begin{table}[htbp]
\centering
\caption{LIFT architecture and training parameters. Widths specify
hidden layers in order. The real-data column applies to both
lineage-traced scRNA-seq applications. ReLU activations and Adam
optimization are used throughout.}
\label{tab:sm_implementation}
\small
\setlength{\tabcolsep}{4pt}
\renewcommand{\arraystretch}{1.12}
\begin{tabular}{lcccc}
\toprule
Parameter & Synthetic  & Real data\\
\midrule
Encoder widths
& $32,64,32$
& $32,128,128,32$\\
Prior widths
& $32,64,32$
& $32,128,128,32$\\
Decoder widths
& $32,64,32$
& $32,64,128,64,32$\\
Latent dimension $d_z$ & 2 & 2\\
Regularization weight $\beta$ & 10  & 2\\
Reconstruction weight $C$ & 1 & 10\\
Learning rate
& $10^{-3}$  & $10^{-4}$\\
Batch size & 1,000 & 1,024\\
Training iterations & 5,000 & 20,000\\
\bottomrule
\end{tabular}
\end{table}

Training pairs are sampled from the prescribed coupling $\pi$, and latent variables are sampled from $q_\phi(\cdot\mid x,y)$ using reparameterization. The time variables are sampled as $t\sim U[0,1]$ and $r\mid t\sim U[0,t]$. The directional derivative in the Mean Flow target is computed by a Jacobian--vector product, and the target is detached during parameter updates. The KL divergences between diagonal Gaussians are evaluated analytically. These operations follow the training algorithm in the main article.

At inference, the encoder is not used. Given an initial state $x$, we
sample $Z\sim p_\eta(\cdot\mid x)$ and evaluate
\[
\widehat X_t=x+t\,u_\theta(x,Z;0,t).
\]
The same latent realization is retained across requested times along
one generated path. Fate probabilities are estimated from 1,000 independent
latent samples per initial state. The real-data experiments use the
two-dimensional SPRING representations described in the main article.
Barcode information is incorporated through the coupling rather than
as an input to the sampling network. Timing measurements are obtained
on an Intel i7-13700 CPU.

\subsection{Hyperparameter sensitivity}
\label{subsec:sm_sensitivity}

We vary the latent dimension $d_z$, reconstruction weight $C$, and standard-Gaussian regularization weight $\beta$ on the synthetic branching benchmark. One parameter is varied at a time, with the others fixed at the default values $(d_z,C,\beta)=(2,1,10)$. Table~\ref{tab:sm_sensitivity} gives the complete results for $\alpha\in\{0.5,1,2\}$.

\begin{table}[htbp]
\centering
\caption{Hyperparameter sensitivity on the synthetic branching
benchmark. The default setting is $(d_z,C,\beta)=(2,1,10)$.
Lower values are better for $W_2$ and PE.}
\label{tab:sm_sensitivity}
\small
\setlength{\tabcolsep}{4pt}
\begin{tabular}{llcccccc}
\toprule
& & \multicolumn{2}{c}{$\alpha=0.5$}
    & \multicolumn{2}{c}{$\alpha=1$}
    & \multicolumn{2}{c}{$\alpha=2$}\\
\cmidrule(lr){3-4}
\cmidrule(lr){5-6}
\cmidrule(lr){7-8}
Parameter & Value
& $W_2$ & PE & $W_2$ & PE & $W_2$ & PE\\
\midrule
Latent dimension $d_z$
& 1 & 0.417 & 0.135 & 0.288 & 0.577 & 0.302 & 1.212\\
& 2 & 0.266 & 0.121 & 0.268 & 0.457 & 0.280 & 0.809\\
& 4 & 0.277 & 0.122 & 0.306 & 0.334 & 0.379 & 1.095\\
\midrule
Reconstruction weight $C$
& 0.5 & 0.250 & 0.179 & 0.279 & 0.537 & 0.261 & 0.936\\
& 1   & 0.266 & 0.121 & 0.268 & 0.457 & 0.280 & 0.809\\
& 2   & 0.272 & 0.199 & 0.286 & 0.446 & 0.279 & 0.569\\
& 4   & 0.290 & 0.207 & 0.335 & 0.431 & 0.367 & 0.670\\
\midrule
Regularization weight $\beta$
& 1  & 0.367 & 0.362 & 0.294 & 0.538 & 0.309 & 1.175\\
& 3  & 0.259 & 0.198 & 0.303 & 0.367 & 0.279 & 0.840\\
& 5  & 0.257 & 0.216 & 0.286 & 0.382 & 0.275 & 0.936\\
& 10 & 0.266 & 0.121 & 0.268 & 0.457 & 0.280 & 0.809\\
\bottomrule
\end{tabular}
\end{table}

These results support the default configuration as a balanced choice across the tested regimes, rather than a uniformly optimal setting.

\subsection{Ablation on LIFT components}
\label{subsec:sm_ablation_components}
\paragraph{Separating representation from decoding}
We next keep the encoder and
conditional prior and change the decoder: a static latent MLP,
a latent FM decoder, or the latent Mean Flow decoder. The FM decoder
in this ablation includes $Z$ and should not be confused with the
unaugmented CFM baseline in Table~\ref{tab:synthetic_results}.

\begin{table}[htbp]
\centering
\caption{Latent-decoder ablation. FM uses 100 Euler steps.
Times are milliseconds per batch of 1,000 points on the reported CPU,
averaged over 100 timing measurements. All three variants retain
latent conditioning.}
\label{tab:decoder_ablation}
\small
\setlength{\tabcolsep}{4pt}
\begin{tabular}{lccccccc}
\toprule
& &\multicolumn{2}{c}{$\alpha=0.5$}
&\multicolumn{2}{c}{$\alpha=1$}
&\multicolumn{2}{c}{$\alpha=2$}\\
\cmidrule(lr){3-4}\cmidrule(lr){5-6}\cmidrule(lr){7-8}
Decoder&Time&$W_2$&PE&$W_2$&PE&$W_2$&PE\\
\midrule
Latent FM &43.16&0.201&0.928&0.234&0.805&0.267&1.714\\
Latent MLP &0.8516&0.280&0.350&0.413&0.813&0.379&1.325\\
Latent Mean Flow &0.9607&0.259&0.198&0.268&0.457&0.280&0.809\\
\bottomrule
\end{tabular}
\end{table}

The results in Table~\ref{tab:decoder_ablation} show a trade-off rather than uniform dominance. Latent FM gives the lowest population $W_2$ in all three settings, while latent Mean Flow gives the lowest PE and avoids iterative integration. The static MLP is slightly faster than Mean Flow but has larger probability errors. Thus, the advantage of the proposed decoder is not that one-step generation is unique to Mean Flow, but that it combines low sampling cost with accurate conditional fate prediction in these experiments.

\subsection{Additional qualitative comparisons}
\label{subsec:sm_visual}

\begin{figure}[htbp]
\centering
\centering
    \begin{subfigure}[b]{0.32\linewidth}
        \centering
        \includegraphics[width=\linewidth]{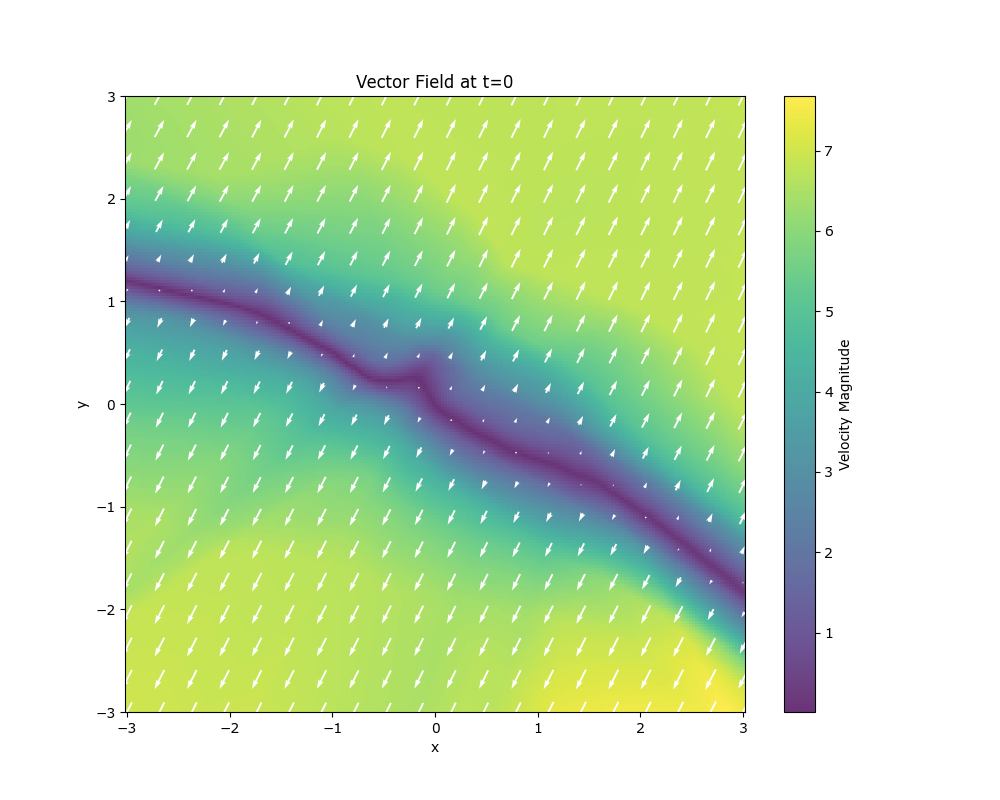}
        \caption{}
        \label{fmv:sub1}
    \end{subfigure}
    \hfill
    \begin{subfigure}[b]{0.32\linewidth}
        \centering
        \includegraphics[width=\linewidth]{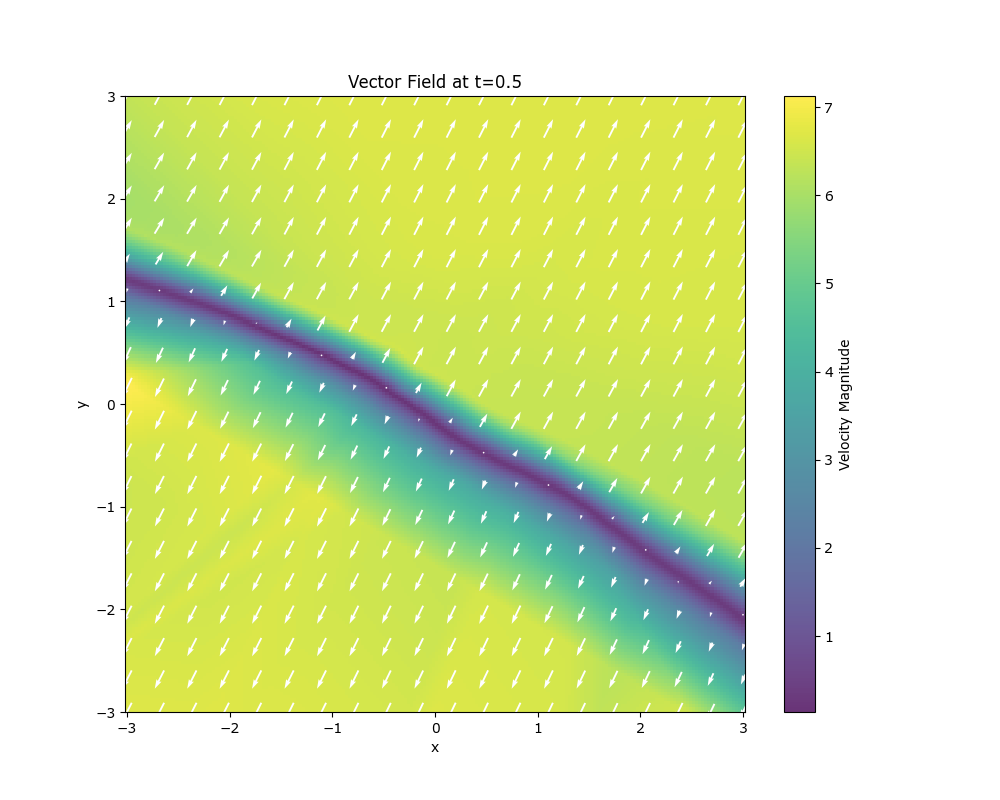}
        \caption{}
        \label{fmv:sub2}
    \end{subfigure}
    \hfill
    \begin{subfigure}[b]{0.32\linewidth}
        \centering
        \includegraphics[width=\linewidth]{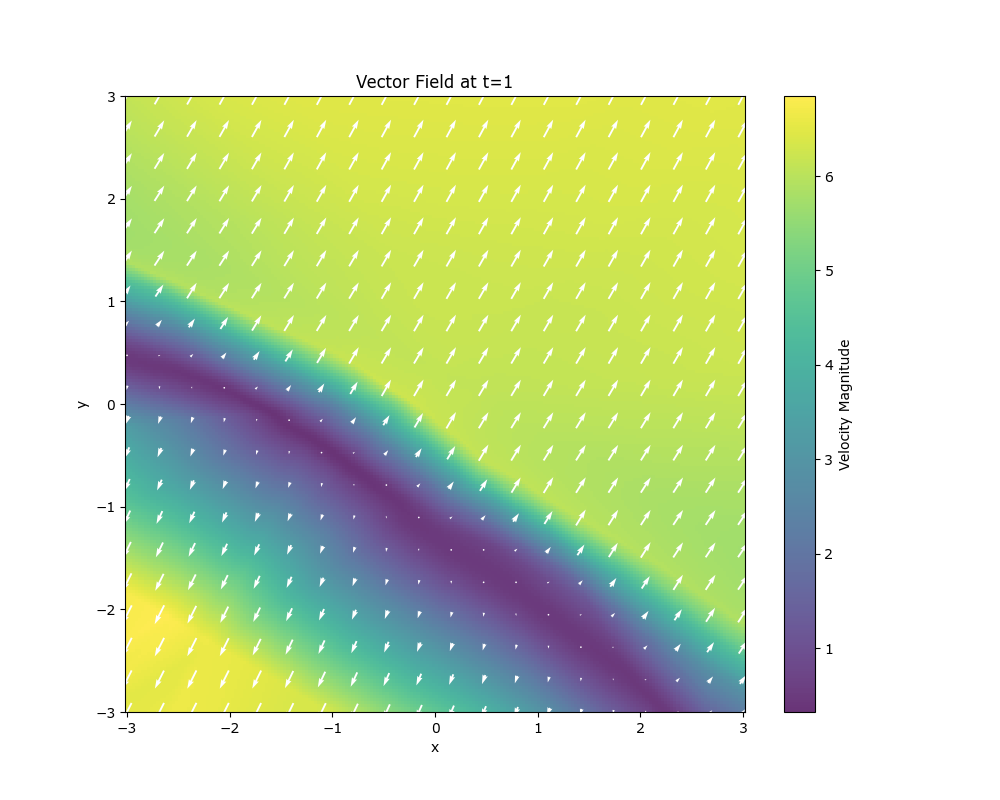}
        \caption{}
        \label{fmv:sub3}
    \end{subfigure}

    \begin{subfigure}[b]{0.32\linewidth}
        \centering
        \includegraphics[width=\linewidth]{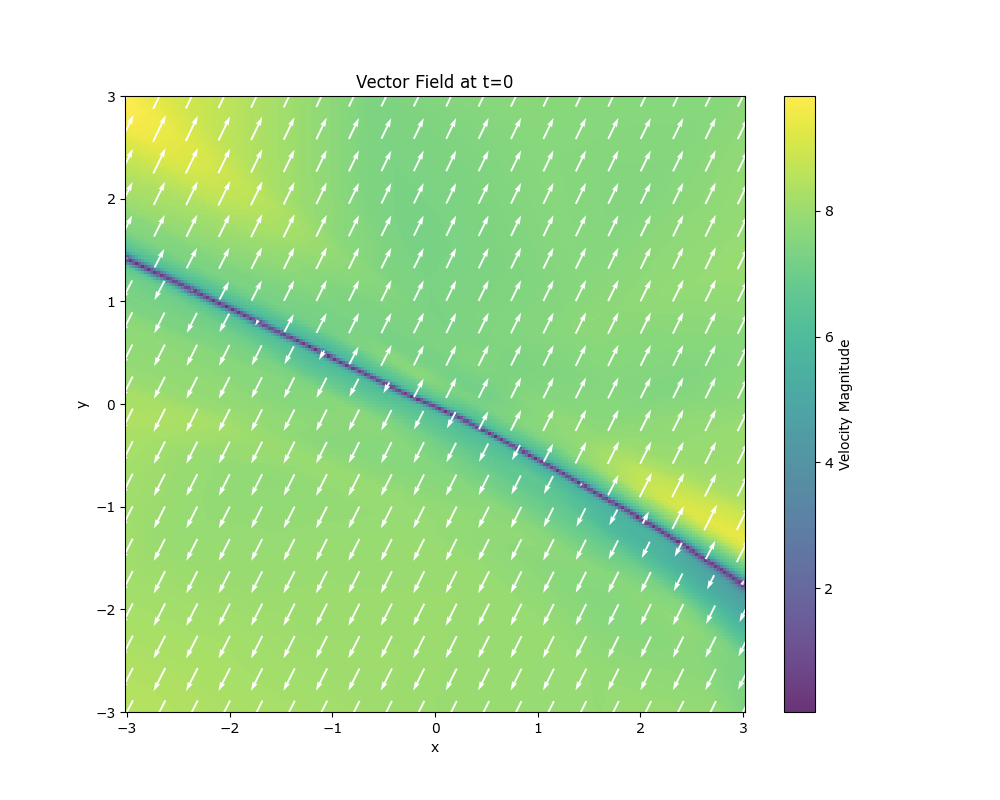}
        \caption{}
        \label{mfv:sub1}
    \end{subfigure}
    \hfill
    \begin{subfigure}[b]{0.32\linewidth}
        \centering
        \includegraphics[width=\linewidth]{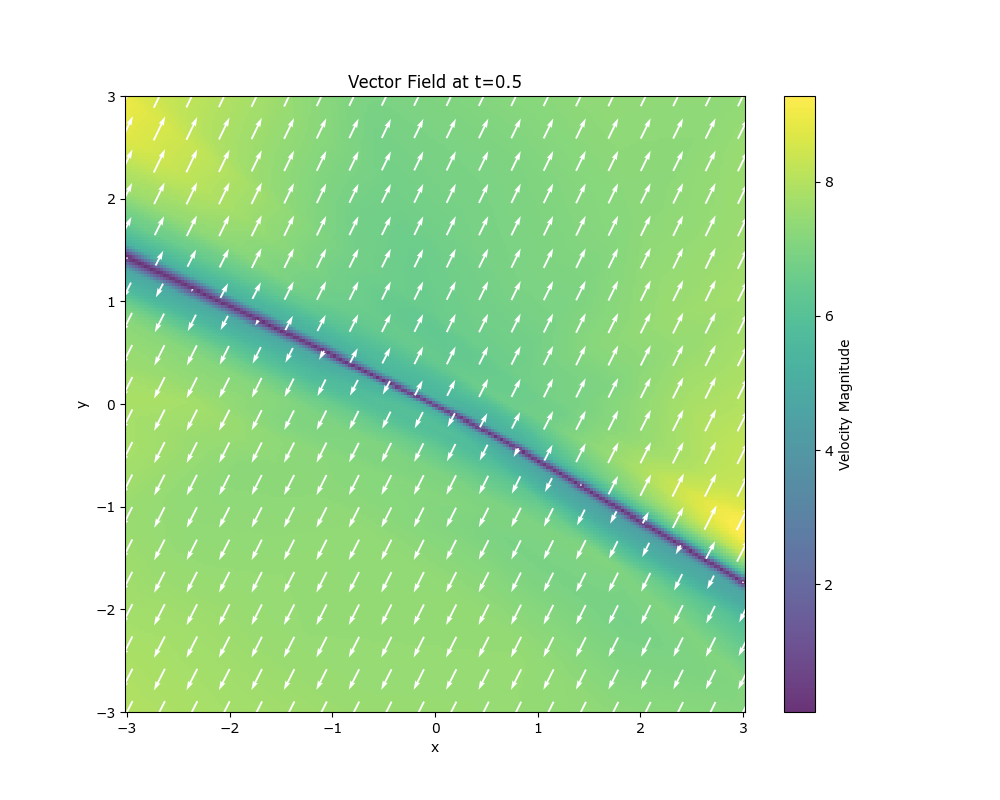}
        \caption{}
        \label{mfv:sub2}
    \end{subfigure}
    \hfill
    \begin{subfigure}[b]{0.32\linewidth}
        \centering
        \includegraphics[width=\linewidth]{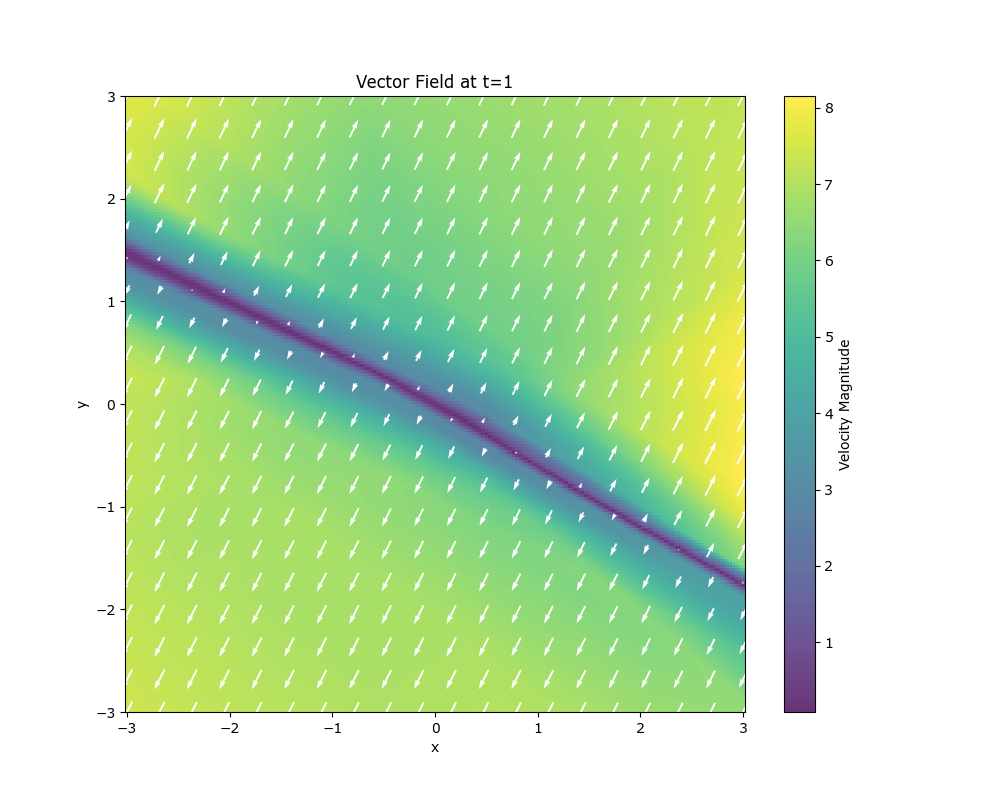}
        \caption{}
        \label{mfv:sub3}
    \end{subfigure}

    \caption{Vector fields derived by standard conditional flow matching (a-c) and mean vector fields derived by standard mean flow (d-f). }

\label{fig:sm_baseline_fields}
\end{figure}

\begin{figure}[htbp]
\centering
    \begin{subfigure}[b]{0.32\linewidth}
        \centering
        \includegraphics[width=\linewidth]{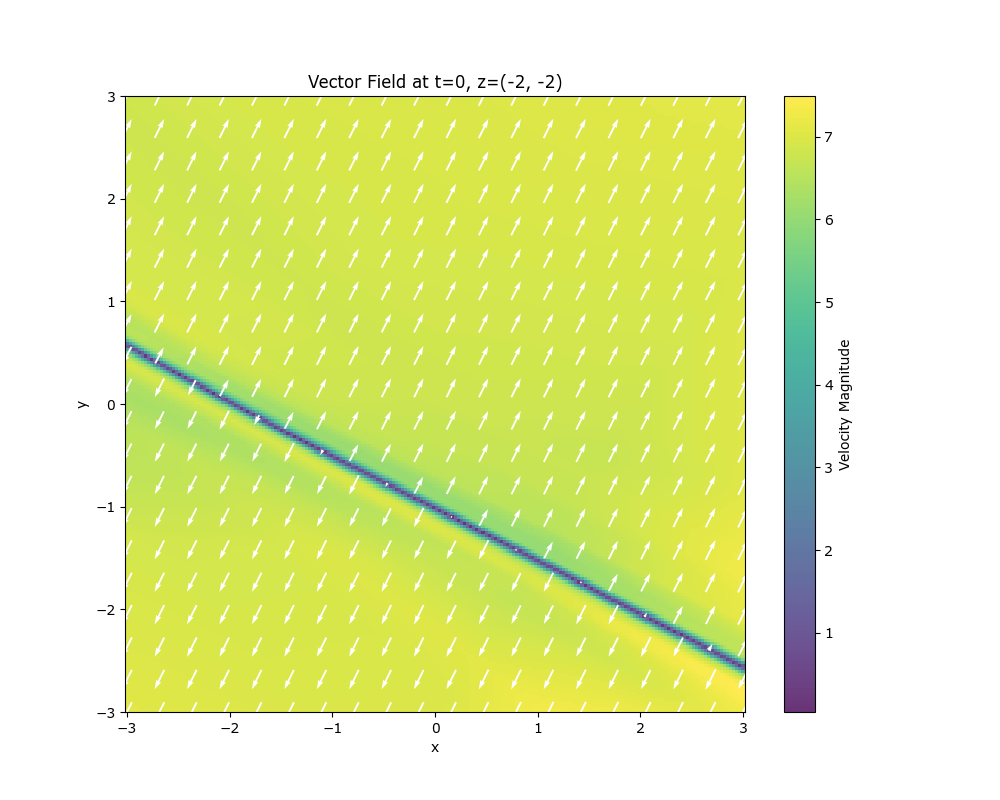}
        \caption{}
        \label{liftv:sub1}
    \end{subfigure}
    \hfill
    \begin{subfigure}[b]{0.32\linewidth}
        \centering
        \includegraphics[width=\linewidth]{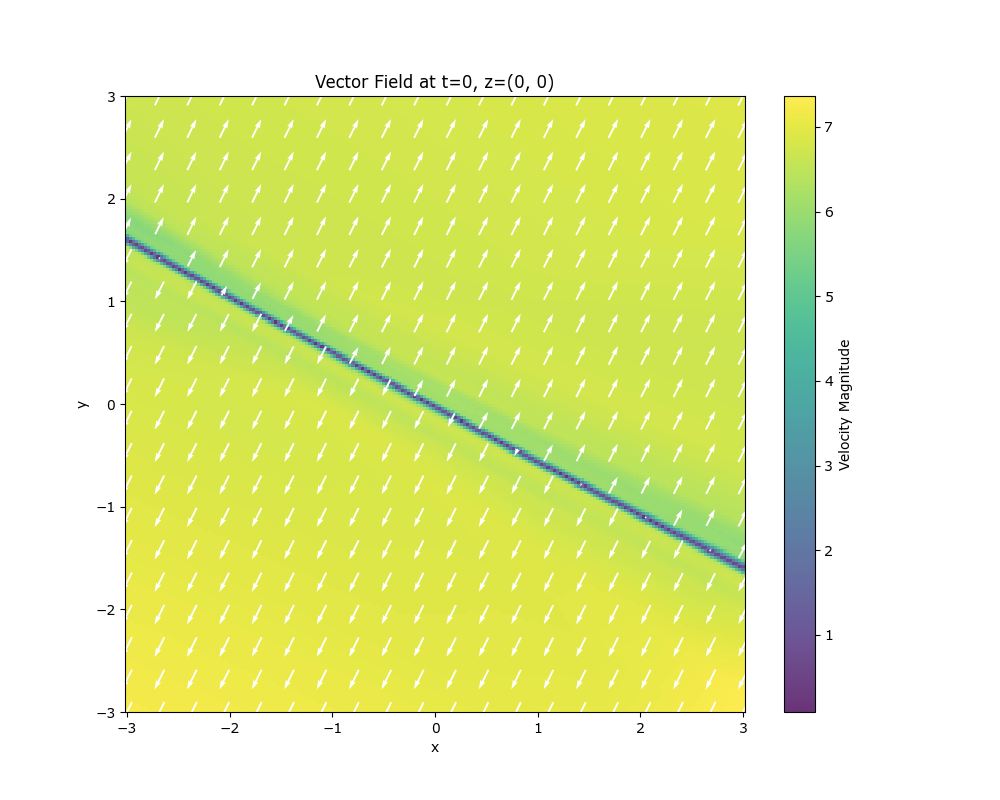}
        \caption{}
        \label{liftv:sub2}
    \end{subfigure}
    \hfill
    \begin{subfigure}[b]{0.32\linewidth}
        \centering
        \includegraphics[width=\linewidth]{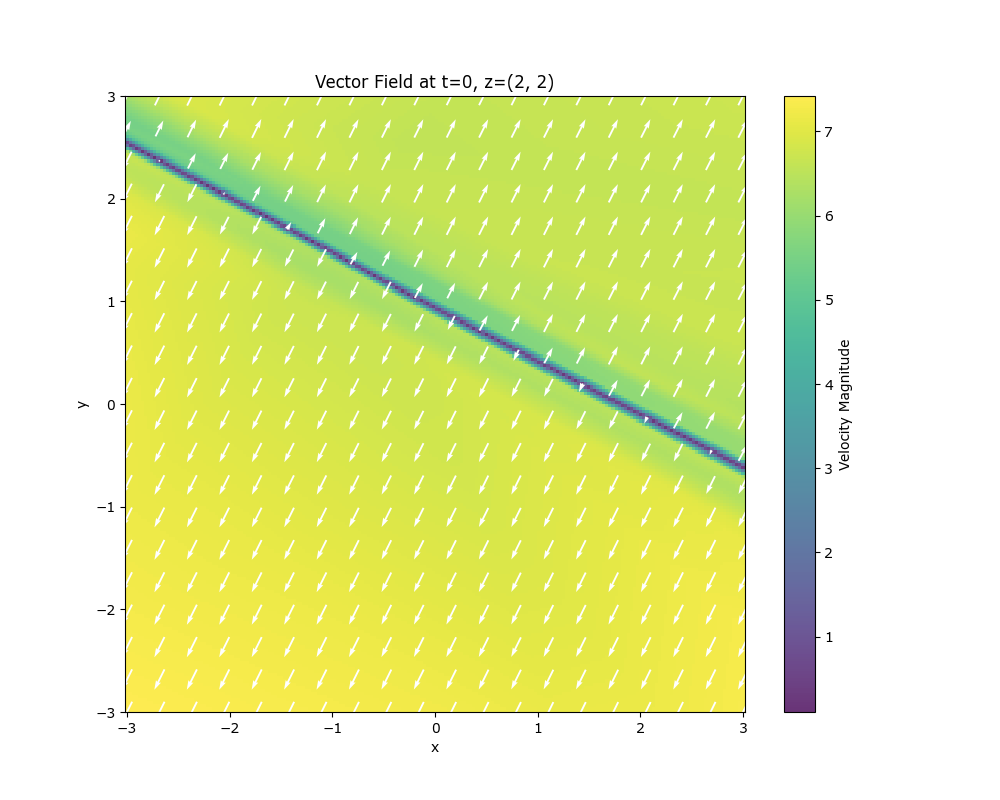}
        \caption{}
        \label{liftv:sub3}
    \end{subfigure}

    \caption{Vector fields derived by LIFT with different latent variable $z$. }
\label{fig:sm_latent_fields}
\end{figure}

\paragraph{Conditioned prediction fields}
Figure~\ref{fig:sm_baseline_fields} displays the instantaneous
velocity fields of CFM and the interval-average velocity outputs
of the unaugmented Mean Flow model. Both exhibit a low-magnitude
region separating the two branch-directed regimes. In
Figure~\ref{fig:sm_latent_fields}, the location of this separating
region changes with the latent code in LIFT. For a fixed code, the
prediction remains deterministic, but different codes allow the
same observed state to receive different predictions. This provides
a qualitative illustration of latent conditioning; it is not a
proof of nonintersection or of a topological property of the
learned dynamics.

\begin{figure}[htbp]
\centering
    \begin{subfigure}[b]{0.4\linewidth}
        \centering
        \includegraphics[width=\linewidth]{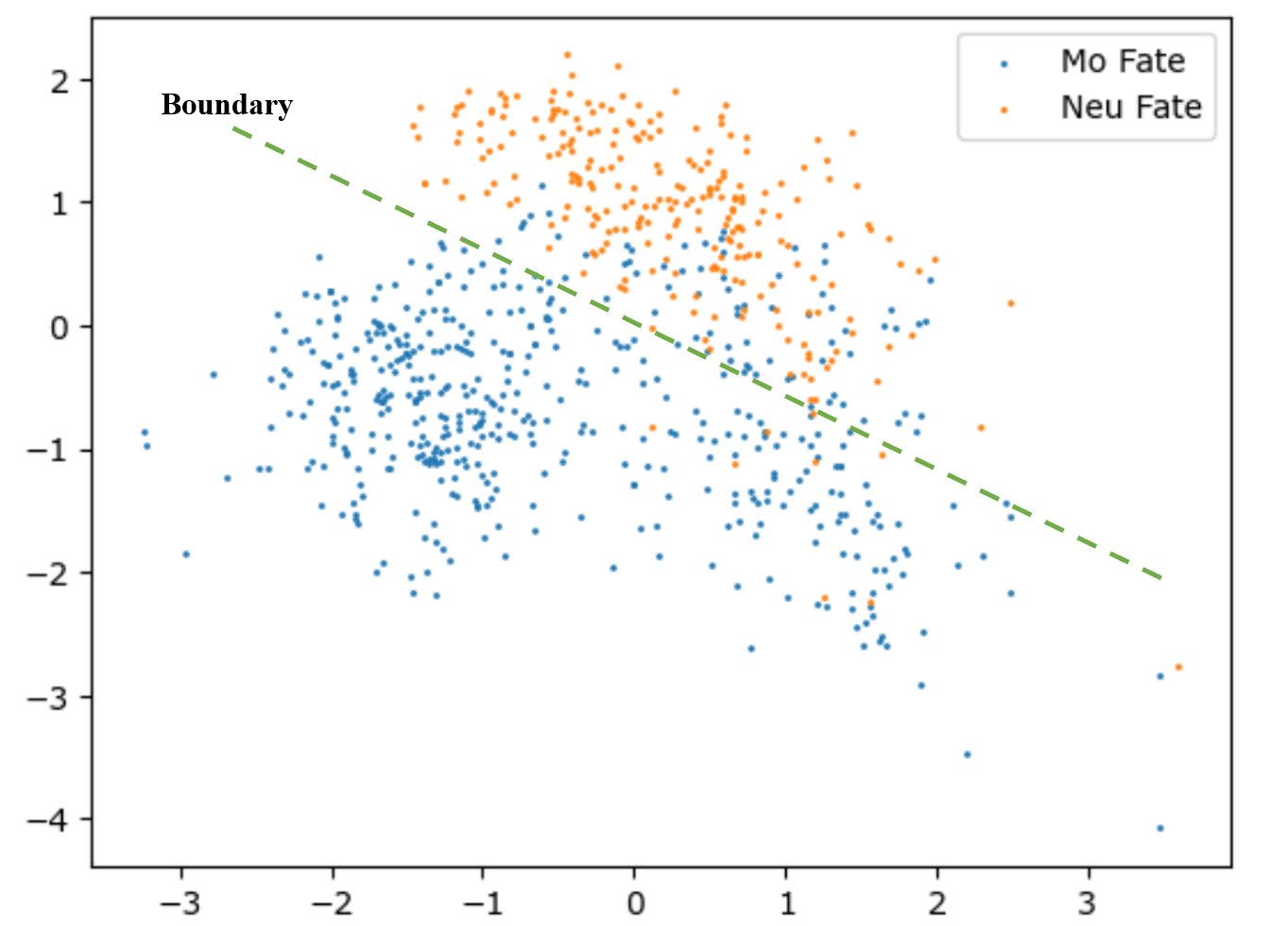}
        \caption{}
        \label{latent:sub1}
    \end{subfigure}
    \begin{subfigure}[b]{0.4\linewidth}
        \centering
        \includegraphics[width=\linewidth]{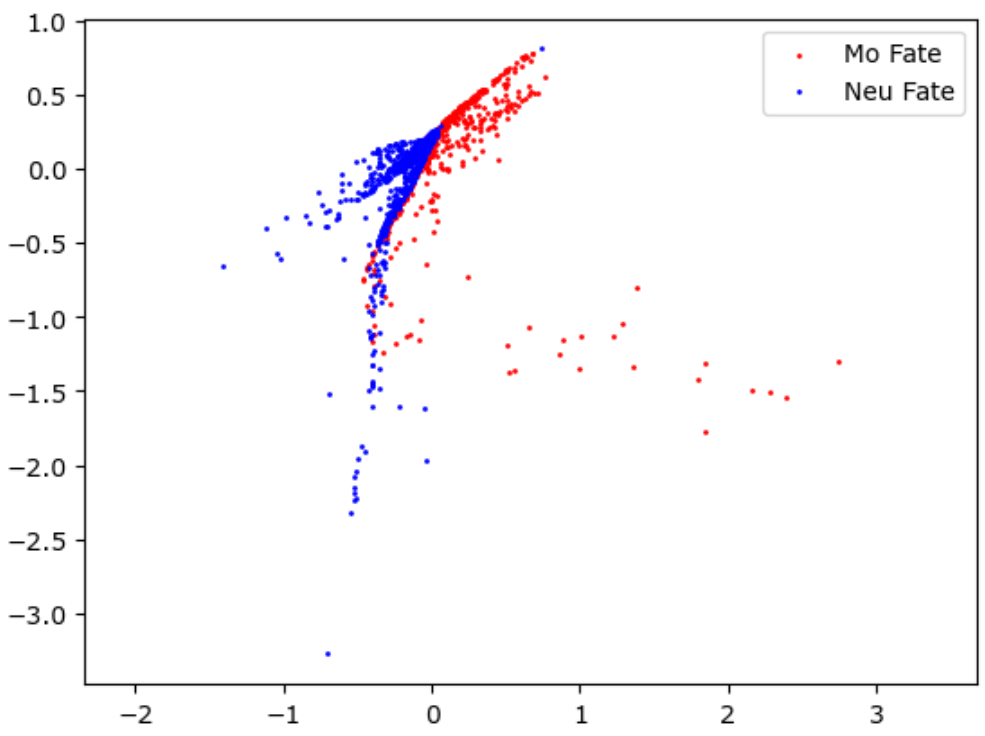}
        \caption{}
        \label{latent:sub2}
    \end{subfigure}
\caption{Latent representations in the hematopoietic experiment.
(a) Posterior means $m_\phi(x,y)$, which use both endpoints.
(b) Conditional-prior means $m_\eta(x)$, which use the initial
state only. Colors indicate the Mo and Neu fate annotations
associated with the displayed observations.}
\label{fig:sm_latent_embeddings}
\end{figure}

\paragraph{Posterior and conditional-prior representations}
Figure~\ref{fig:sm_latent_embeddings} compares posterior means
$m_\phi(x,y)$ and conditional-prior means $m_\eta(x)$ for the
hematopoietic application. Endpoint pairs associated with Mo
and Neu outcomes occupy different regions in the posterior
representation. The encoder receives the terminal state, so
this separation illustrates how endpoint information is encoded,
rather than an independent prediction made from the initial state.
The prior representation uses the initial state alone and also
shows fate-associated structure. Fate probabilities, however,
are obtained by sampling the full prior and applying the decoder,
not by assigning a fate to the prior mean. These visualizations
characterize learned representations without identifying a latent
biological mechanism.

\begin{figure}[htbp]
\centering
\centering
    \begin{subfigure}[b]{0,32\linewidth}
        \centering
        \includegraphics[width=\linewidth]{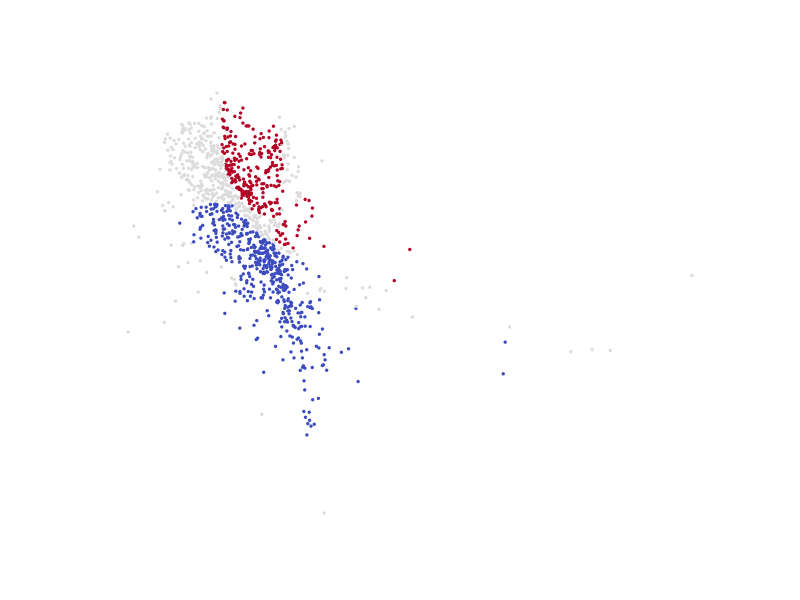}
        \caption{MF-Geo}
        \label{other_he_sub1}
    \end{subfigure}
    \hfill
    \begin{subfigure}[b]{0,32\linewidth}
        \centering
        \includegraphics[width=\linewidth]{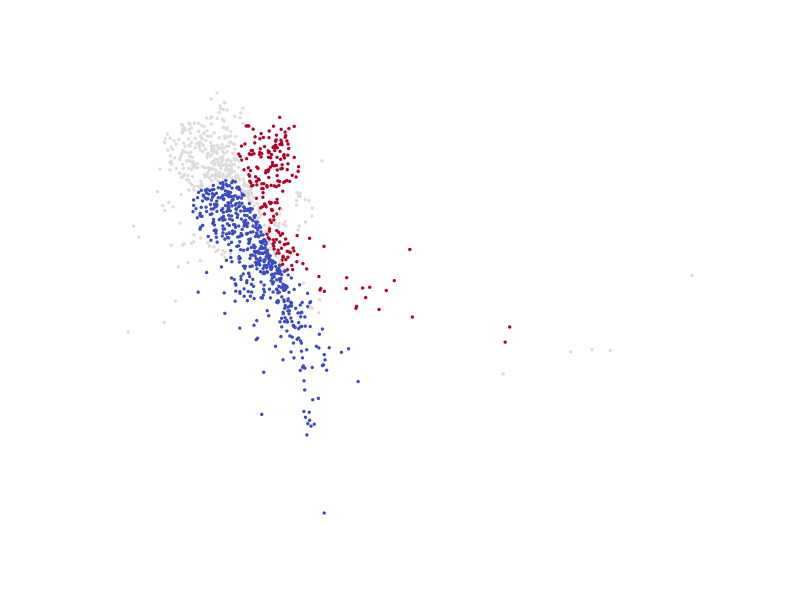}
        \caption{MF-Lin}
        \label{other_he_sub2}
    \end{subfigure}
    \hfill
    \begin{subfigure}[b]{0,32\linewidth}
        \centering
        \includegraphics[width=\linewidth]{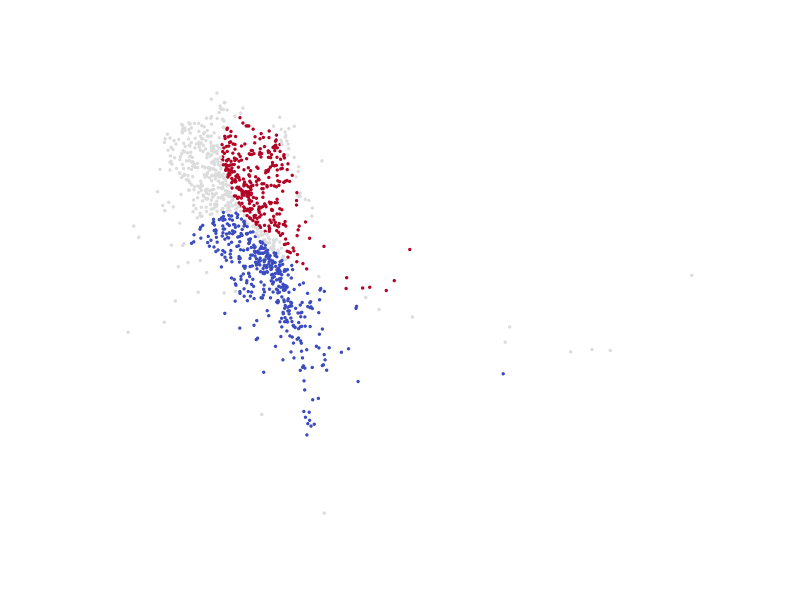}
        \caption{CFM-Geo}
        \label{other_he_sub3}
    \end{subfigure}

    \begin{subfigure}[b]{0,32\linewidth}
        \centering
        \includegraphics[width=\linewidth]{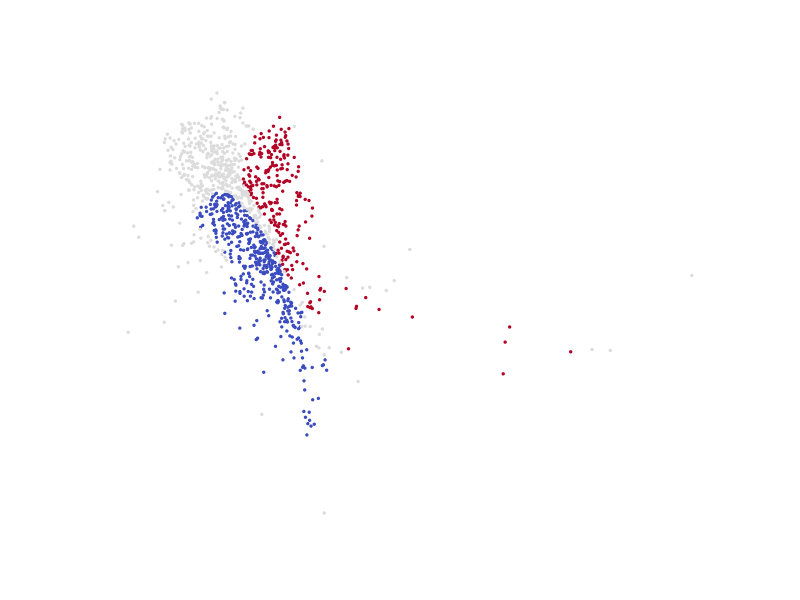}
        \caption{CFM-Lin}
        \label{other_he_sub4}
    \end{subfigure}
    \hfill
    \begin{subfigure}[b]{0,32\linewidth}
        \centering
        \includegraphics[width=\linewidth]{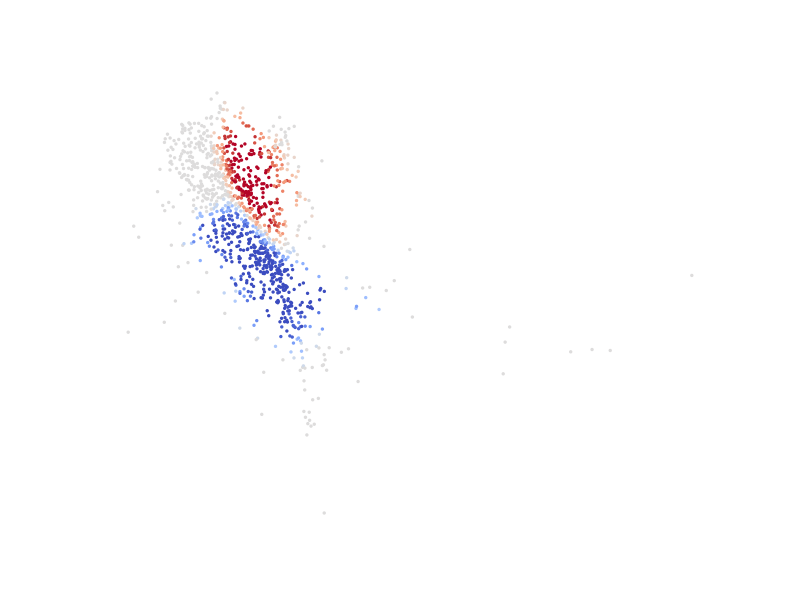}
        \caption{SM}
        \label{other_he_sub6}
    \end{subfigure}

    \caption{Visualization of cell fate probabilities for other baseline methods on the hematopoiesis dataset. Colors match the main text. }
\label{fig:sm_hematopoiesis_maps}
\end{figure}

\paragraph{Baseline fate maps}
Figures~\ref{fig:sm_hematopoiesis_maps} provide the complete baseline
fate maps for the hematopoiesis applications. The maps compare
geometric and lineage-informed couplings for CFM and Mean Flow,
together with SM. They complement the numerical
population and probability scores by showing where the spatial
patterns of predicted fates differ. Gray regions in the baseline
visualizations mark predictions outside the assigned fate regions,
rather than an additional biological fate. The maps should be read
together with the quantitative comparisons, since visually separated
outcomes do not by themselves establish accurate conditional
probabilities or irreversible fate commitment.

\bibliographystyle{siamplain}
\bibliography{references}

\end{document}